\documentclass[lettersize,journal]{IEEEtran}
\usepackage{amsmath,amsfonts}
\usepackage{algorithmic}
\usepackage{array}
\usepackage[caption=false,font=normalsize,labelfont=sf,textfont=sf]{subfig}
\usepackage{textcomp}
\usepackage{stfloats}
\usepackage{url}
\usepackage{verbatim}
\usepackage{graphicx}
\def\BibTeX{{\rm B\kern-.05em{\sc i\kern-.025em b}\kern-.08em
    T\kern-.1667em\lower.7ex\hbox{E}\kern-.125emX}}
\usepackage{balance}

\usepackage{amsthm}
\usepackage{amssymb}
\usepackage{booktabs}
\usepackage{algorithm}
\usepackage{color}
\usepackage{multirow}

\begin{document}

\title{Enhancing Unsupervised Anomaly Detection \\ with Score-Guided Network}

\author{Zongyuan Huang,~Baohua Zhang,~Guoqiang Hu,~Longyuan Li,~Yanyan Xu{$^\ast$},~Yaohui Jin{$^\ast$},~\IEEEmembership{Member,~IEEE}
\IEEEcompsocitemizethanks{
\IEEEcompsocthanksitem Z. Huang, L. Li, Y. Xu, Y. Jin are with the MoE Key Laboratory of Artificial Intelligence and AI Institute, Shanghai Jiao Tong University, Shanghai 200240, China. Z. Huang is also with Shanghai Key Laboratory of Multidimensional Information Processing, East China Normal University, Shanghai 200241, China. E-mail: \{herozen, jeffli, yanyanxu, jinyh\}@sjtu.edu.cn.
\IEEEcompsocthanksitem B. Zhang and G. Hu are with the Big Data and AI Lab of Industrial and Commercial Bank of China, Shanghai 201206, China. E-mail: \{zhangbh, hugq\}@sdc.icbc.com.cn\\
$^\ast${}Corresponding \ authors.
}
}


\maketitle

\begin{abstract}
Anomaly detection plays a crucial role in various real-world applications, including healthcare and finance systems. Owing to the limited number of anomaly labels in these complex systems, unsupervised anomaly detection methods have attracted great attention in recent years. Two major challenges faced by existing unsupervised methods are: (i) distinguishing between normal and abnormal data 
when they are highly mixed together; (ii) defining an effective metric to maximize the gap between normal and abnormal data in a hypothesis space, which is built by a representation learner.
To that end, this work proposes a novel scoring network with a score-guided regularization to learn and enlarge the anomaly score disparities between normal and abnormal data, enhancing the capability of anomaly detection. With such score-guided strategy, the representation learner can gradually learn more informative representation during the model training stage, especially for the samples in the transition field. 
Moreover, the scoring network can be incorporated into most of the deep unsupervised representation learning-based anomaly detection models and enhances them as a plug-in component.  
We next integrate the scoring network into an autoencoder and four state-of-the-art models to demonstrate the effectiveness and transferability of the design. These score-guided models are collectively called SG-Models.
Extensive experiments on both synthetic and real-world datasets confirm the state-of-the-art performance of SG-Models.
\end{abstract}


\section{Introduction}
\label{sec:introduction}

\IEEEPARstart {A}{nomaly} detection aims to identify the samples that considerably deviate from the expectation of a complex system. There are plentiful applications of anomaly detection across various domains~\cite{survey2019,surveyPang,li2020anomaly}, for example, disease detection~\cite{wong2002rule}, network intrusion detection~\cite{falcao2019quantitative}, and financial fraud detection~\cite{ahmed2017anomaly}, etc. Existing anomaly detection methods can be grouped into three categories according to the label availability in the training datasets: supervised learning methods, semi-supervised learning methods, and unsupervised learning methods~\cite{han2022adbench}. Supervised and semi-supervised learning methods rely on labeled data, which rarely exist in some practical applications. Accordingly, unsupervised anomaly detection methods have received very much attention in both academic and industrial communities.

\begin{figure*}[bt]
        \begin{center}
        \includegraphics[width=0.88\textwidth]{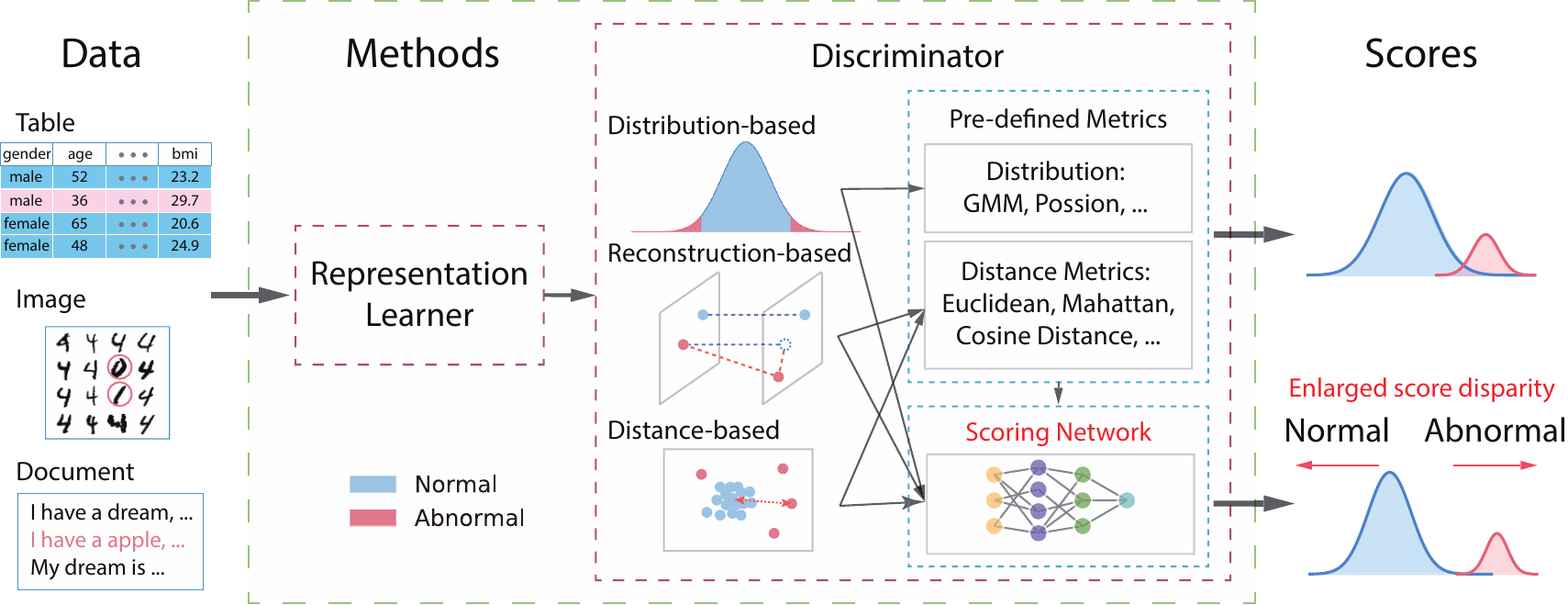}
        \end{center}
        \caption{A typical pipeline of unsupervised representation learning-based anomaly detection method and the proposed scoring network. The scoring network can be embedded into existing methods without additional assumptions. Using the pre-defined metric as a self-supervised signal, the scoring network distinguishes the obvious-normal and obvious-abnormal samples and makes full use of them to strengthen the representation learner and to enlarge the score disparity between normal and abnormal samples, enhancing the discrimination ability of the original method.}
        \label{fig:framework}
    \end{figure*}

    In general, unsupervised methods assume that most of the data are normal samples and expect to fit the pattern of normal samples to distinguish anomalies. A mainstream unsupervised anomaly detection methodology is based on representation learning and the typical pipeline is presented in Figure~\ref{fig:framework}. 
    A representation learner is first trained to map the original data, e.g., tabular, image, or document data, into the desired hypothesis space. Then a discriminator mostly uses various pre-defined metrics to identify the abnormal samples. Based on the definition of the discriminator, we can further classify the unsupervised anomaly detection methods into three categories: distribution-based methods, distance-based methods, and reconstruction-based methods~\cite{surveyUnify}. The distribution-based methods identify anomalies by assuming the normal data follow a certain distribution; the distance-based methods attempt to enhance the distance between the abnormal and normal data in a latent space; the reconstruction-based methods identify anomalies by the similarity between original data and reconstructed data. The final identification of anomaly always solely relies on the pre-defined distance calculated between the original samples or their representations in these existing methods, as illustrated in Figure~\ref{fig:framework}.
    
\begin{figure}[bt]
        \centering
        \includegraphics[width=0.45\textwidth]{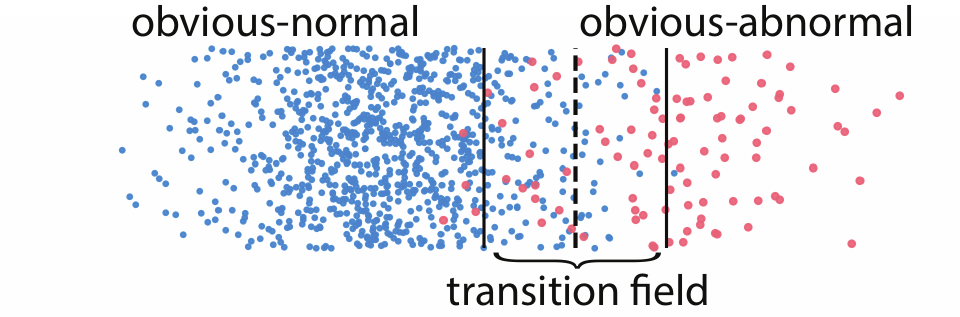}
        \caption{A toy example to illustrate the concepts of obvious-normal, obvious-abnormal, and transition fields.}
        \label{fig:illustration}
    \end{figure}

    Most existing unsupervised methods (including some of semi-supervised methods) face the following challenges: 
    (1) When handling training datasets comprising both normal and abnormal samples, one typical strategy is to fit all data to a normal pattern and identify samples with a weak fit as anomalies~\cite{golan2018deep,ruff2018deep,dagmm,goyal2020drocc,goad,shenkar2022anomaly}. An alternative strategy is to filter suspected anomalies and fit the remaining samples during the training process~\cite{rda,rdp,rsrae}. However, these strategies may not effectively deal with the anomaly-contained situation, where we can utilize the information of suspected anomalies to enhance the model's detection capability.
    (2) The manually selected metric to compute the anomaly scores will impact the performance and can hardly be "one-model-fits-all"~\cite{rda,dagmm,rsrae}. 
    (3) Data distribution can be divided into three fields: obvious-normal field, transition field, and obvious-abnormal field, as shown in Figure~\ref{fig:illustration}. The samples in obvious-normal or obvious-abnormal fields can be clearly distinguished as normal or abnormal data. However, the mixture of normal and abnormal data in the transition field hampers the identification of the abnormal samples.

    To overcome these challenges, we propose a scoring network to promote the performance of anomaly identification via learning and enhancing the disparity between normal and abnormal data. As shown in Figure~\ref{fig:framework}, the scoring network can be incorporated into an existing deep unsupervised representation learning-based anomaly detection method and directly connected with the representation learner, which maps the original data into a hypothesis space.
    The scoring network utilizes the original pre-defined metric as a self-supervised signal to distinguish obvious-normal and obvious-abnormal samples and leverages these samples to guide the training of the representation learner to extract valuable information, which is more helpful for discrimination.
    Specifically, we introduce a score-guided regularization in the scoring network to gradually intensify the divergence between obvious-normal and obvious-abnormal samples during model training. We assign smaller scores to the obvious-normal data and larger scores to the obvious-abnormal data. Paying attention to the distinguishable samples accelerates the model's convergence and enhances the capability of both the representation learner and the scoring network. 
    This strategy leads to an incremental increase in score disparity between normal and abnormal data and facilitates the discrimination of non-obvious data in the transition field throughout the model training process.
    Moreover, the anomaly scores are calculated by the discrimination metric learned by the scoring network rather than the pre-defined one, enhancing the transferability of the original method.
    Without additional assumptions, this scoring network can be flexibly embedded into most existing anomaly detection models.

    We further propose a novel instantiation, embedding the scoring network into an autoencoder framework, named score-guided autoencoder (SG-AE). The autoencoder reconstructs data instances and learns their representations in a latent space. Then, the scoring network assigns scores to these representations and attempts to maximize the disparity between the normal and abnormal samples. A sample with a higher anomaly score is more likely to be an anomaly. 
    To explain how the scoring network works, we conduct simulation experiments to observe the changes in the score differences in the transition field. Also, we compare the performances of SG-AE with classic and state-of-the-art methods on seven tabular datasets. We also present the distributions of anomaly scores of normal and abnormal data and utilize t-SNE to reveal the anomaly detection performance in a 2D space. Moreover, we extend the scoring network to three state-of-the-art models and explore the performance improvement of these score-guided models (SG-Models). We also examine the potential of using the scoring network on two document datasets and one image dataset. The main contributions of this work are summarized as follows:
    
    \begin{itemize}
    \item We propose a scoring network with a score-guided regularization, which can directly learn the anomaly scores and be easily integrated into most unsupervised representation learning-based models without imposing more assumptions. The scoring network utilizes the obvious data, distinguished by the self-supervised signal, to strengthen the representation learner and expand the learned scores disparity between normal and abnormal samples in the transition field, thus enhancing the anomaly detection ability.
    \item We integrate the scoring network into an autoencoder structure and introduce a simple but effective instantiation, SG-AE. SG-AE breaks the limitation of the autoencoder on normal data input and achieves competitive performance on several datasets.
    We also incorporate the scoring network into four state-of-the-art models to examine the adaptability to different methods. 
    \item Extensive experiments on synthetic and real-world datasets demonstrate the effectiveness of the scoring network. 
    The SG-AE outperforms the classic and state-of-the-art methods on tabular datasets and the other SG-Models show performance improvements with the help of a scoring network. Moreover, our scoring network is robust to the anomaly rate by learning the disparity between normal and abnormal data, which is also demonstrated by the experiments.
    \end{itemize}

\section{Related Work}
\label{sec:related}

    In this section, we briefly review the related work in unsupervised representation learning-based (URL) anomaly detection and revisit the use of regularization techniques in anomaly detection. Then we compare our design with the existing methods.

\subsection{URL Anomaly Detection}

    \textbf{Distribution-based methods}. Distribution-based methods assume that normal data follow a certain distribution in the original or latent space and anomalies deviate from the distribution~\cite{eskin2000anomaly,kde,ebm}. The classic distribution-based methods, based on extreme-value analysis, consider the tail of a probability distribution as anomalies~\cite{book}. In recent years, deep learning brings more possibilities with more distribution hypotheses in the hidden space. For instance, Deep Autoencoding Gaussian Mixture Model (DAGMM)~\cite{dagmm} assumes that the low-dimensional representations and reconstruction errors of normal data follow a Gaussian Mixture Model, and uses the probability value to distinguish between normal and abnormal data. 
    Some one-class classification methods also fall into this category due to the consideration of manifold distribution. Deep SVDD~\cite{ruff2018deep} attempts to map the normal data into a hypersphere in a latent space and the abnormal data fall outside the hypersphere. DROCC~\cite{goyal2020drocc} assumes that the normal data lie on a low-dimensional manifold and distinguishes data via Euclidean distance. It generates synthetic abnormal data to learn a robust boundary with a well-designed optimization algorithm. GOAD~\cite{goad} maps data into a sphere feature space with random affine transformations, and anomalies are assumed to deviate from the center. Some methods, like, Outlier Exposure~\cite{oe}, utilize extra information provided by auxiliary datasets to learn the normal distribution and detect the out-of-distribution samples. Note that although a number of generative methods assume various distributions in latent spaces, they finally determine anomalies based on the reconstruction errors. As such, we classify them into reconstruction-based methods.
    
    \textbf{Distance-based methods}. Distance-based methods consider the positional relations or neighbor structures in the original or hidden space, assuming that anomalous samples stay far away from the normal ones~\cite{knn,gu2019statistical}. Nearest-neighbor-based methods, such as LOF~\cite{lof} and DN2~\cite{bergman2020deep}, are typical of distance-based methods. Likewise, the tree-based methods, like Isolation Forest (iForest)~\cite{iforest} and RRCT~\cite{guha2016robust}, can be regarded as distance-based methods since they attempt to capture high-density fields that are reflected in the depth or complexity of the tree~\cite{surveyUnify}. Deep learning technologies are advantageous tools to learn data representation so that distance can be measured in a hypothesis space~\cite{repen}. For example, Random Distance Prediction (RDP)~\cite{rdp} builds deep neural networks to learn a certain random mapping, which can preserve the proximity information for data. 

    \textbf{Reconstruction-based methods}. Reconstruction-based methods train models to reconstruct data, with an assumption that the trained models learn the patterns of the major data and anomalies are unable to be well reconstructed. Autoencoder framework~\cite{ae} is one fundamental architecture in reconstruction-based methods. Robust Deep Autoencoder (RDA)~\cite{rda} is an early work to build an autoencoder on corrupted data. It isolates noise and outliers with robust principal component analysis, and thereafter trains an autoencoder. Robust Subspace Recovery Autoencoder (RSRAE)~\cite{rsrae} constructs a robust subspace recovery layer within an autoencoder with a well-designed loss. Cosine similarity between original and reconstructed data is used to identify the abnormal samples. CoRA~\cite{cora} is a transductive semi-supervised model, which modifies the autoencoder framework with one encoder and two decoders. The two decoders distinguish normal data from abnormal data by reconstruction errors and then decode them separately. Generative models for anomaly detection can also be considered as reconstruction-based methods. VAE adopts reconstruction probability to identify anomalies~\cite{an2015variational,donut,li2020anomaly} while GAN-based methods detect anomalies by the discriminator due to the inaccurate reconstruction by the generator~\cite{schlegl2017unsupervised,gan,li2018anomaly}.

\subsection{Anomaly Detection with Regularization}
    Regularization is a widely used technique to alleviate the over-fitting problem~\cite{srivastava2014dropout} and achieve adversarial robustness~\cite{wen2020towards}.
    Many pieces of work utilize regularization to encourage autoencoder to learn an informative low dimensional representation of input data~\cite{arpit2016regularized,yu2018learning,qian2019improving,xu2020learning}.
    In the context of anomaly detection, regularization methods are introduced to enforce robustness against anomalies, encouraging the model to learn key underlying regularities~\cite{chalapathy2017robust}. Motivated by robust principal component analysis (RPCA), RDA~\cite{rda} adds sparsity penalty $\ell_1$ and group penalty $\ell_{2,1}$ into its RPCA-like loss function of auto-encoder to improve robustness. RSRAE~\cite{rsrae} further incorporates a special regularizer into the embedding of the encoder to enforce anomaly-robust linear structure. DAGMM~\cite{dagmm} introduces $\ell_1$ penalty to constrain training. Some one-class classification methods also employ the regularization to better learn the boundary and improve performance. Ruff et al. presented a unified view of anomaly detection methods and comprehensively summarized the methods using regularizations~\cite{surveyUnify}. 

\subsection{Comparison with existing methods}
    Compared to the existing work, the proposed method has the following characteristics to tackle the challenges mentioned in Section~\ref{sec:introduction}:
    (1) Many existing unsupervised models are only fitting normal samples~\cite{golan2018deep,ruff2018deep,dagmm,goyal2020drocc,goad,shenkar2022anomaly}. Although some methods can deal with the contaminated data~\cite{rda,rdp,rsrae}, they make efforts to retain the obvious-normal data and filter out the suspected abnormal data. Our method leverages both normal and abnormal samples in the datasets and attempts to enhance the detection capabilities in the transition field. Specifically, our score-guided regularization utilizes the obvious data to enlarge the score disparity between normal and abnormal data, and guide the training of the representation learner and the scoring network. By doing so, the proposed scoring network can improve the ability of the representation learner and the robustness of the entire detection method.  
    (2) The existing unsupervised models define anomaly scores with selected appropriate metrics, such as Euclidean distance for distance-based methods~\cite{ruff2018deep,repen,goyal2020drocc} or cosine similarity for reconstruction-based methods~\cite{rda,schlegl2017unsupervised,rsrae}. The anomaly scores are not directly optimized in an end-to-end fashion. Devnet~\cite{devnet} is the first weakly supervised method to achieve end-to-end anomaly score learning. However, Devnet guides model training with labels that are missing in unsupervised settings and treats the unknown anomalies as normal data. In contrast, our scoring network can be incorporated into existing unsupervised methods without additional assumptions, expanding their ability to handle contaminated data sets and directly optimize anomaly scores.      
    
\section{Methodology}
    
    In this section, we first introduce the score-guided regularization and then incorporate the scoring network into URL methods. These score-guided models are collectively called SG-Models. Finally, the algorithm of SG-Models is elaborated.  

\subsection{Score-guided Regularization}
\label{sec:SGR}

    Let $\mathbf{X}=\{\boldsymbol{x}_i\}_N$ denote a dataset with $N$ samples where the i-th sample $\boldsymbol{x}_i$ is a $D$ dimensional vector, $\boldsymbol{x}_i\in\mathbb{R}^D, i\in N$. 
    The common practice of URL anomaly detection methods is to choose one hypothesis mentioned in Section~\ref{sec:related} to build a model. As shown in Figure~\ref{fig:framework}, after the representation learner maps the data samples to a latent space, a pre-defined function $f(\cdot)$ is selected to identify the anomalous data in the latent space. ``$\cdot$'' in $f(\cdot)$ means that the function $f(\cdot)$ has different forms according to the chosen hypothesis. Specifically, $f(\cdot)$ can be distribution functions $f(\boldsymbol{z})$, reconstruction errors $f(\boldsymbol{x}, \boldsymbol{\tilde{x}})$, or distance relations $f(\boldsymbol{z}_i, \boldsymbol{z}_j)$, as:
    \begin{equation}\label{eq:f}
        f(\cdot) = 
        \begin{cases}
        f(\boldsymbol{z}_i), & \text{distribution-based} \\
        f(\boldsymbol{x}_i, \boldsymbol{\tilde{x}}_i), & \text{reconstruction-based} \\
        f(\boldsymbol{z}_i, \boldsymbol{z}_j), & \text{distance-based}
        \end{cases},
    \end{equation}
    where $\boldsymbol{z}_{i(j)}$ is the learned data representation of $\boldsymbol{x}_{i(j)}$, $\boldsymbol{\tilde{x}}_i$ is the reconstructed counterpart of $\boldsymbol{x}_i$.
    
    We can calculate a metric value $f$ of a sample with the function $f(\cdot)$. The metric values of anomalies deviate from the major samples, and a larger deviation suggests a higher probability of being an anomaly. However, selecting or designing an effective $f(\cdot)$ requires prior domain knowledge about the tasks and there is almost no universal metric to fit various datasets. 
    Therefore, an intuitive practice is to directly learn the metric through a neural network, denoted as the scoring network, rather than a pre-defined one. 
    The learned metric can be used to evaluate the abnormal degrees of samples, and it is more reliable because it is learned from the data without assumption bias.  
    We expect that the scores of normal data and abnormal data under this metric are significantly different, thus achieving more accurate anomaly detection.
    However, it seems impossible for the scoring network to assign correct anomaly scores to samples in an unsupervised setting, especially for samples in transition fields.
    
    In fact, we can use the pre-defined metric value as a self-supervised signal to guide the training of the scoring network. 
    We suppose that the metric function has the ability to assign different reference scores to the obvious-normal and obvious-abnormal data. The scoring network assigns anomaly scores to data representations learned by the representation learner and guides anomaly score distribution based on the reference scores.
    Note that the reference scores are calculated by the pre-defined metric function to prejudge the normal and abnormal samples with obvious differences, while the anomaly scores are learned by the network and used to finally evaluate the abnormalities of all samples. 
    The scoring network is expected to guide the anomaly score distribution of the obvious data first based on the reference scores, and then gradually try to guide the non-obvious data in the transition field based on the information learned from obvious data.
    
    To achieve this, we propose a scoring network with a score-guided regularization to learn and enlarge the anomaly score disparity and to enhance the capability of the representation learner, as shown in Figure~\ref{fig:framework}.  
    Supposing $\boldsymbol{z}$ is the representation of a data sample in the latent space, the scoring network is devised to take $\boldsymbol{z}$ as input and outputs an anomaly score in an end-to-end manner. 
    As expected to assign smaller anomaly scores to obvious-normal samples and larger scores to obvious-abnormal samples, the regularization function $L_{\rm{SE}}$ can be defined as follows:
    \begin{equation}\label{eq:SELoss_0}
        L_{\rm{SE}} (f, s) = 
        \begin{cases}
        |s|, & f < \epsilon_1 \\
        \lambda_a \cdot \max(0, a-s), & f > \epsilon_2
        \end{cases},
    \end{equation}
    where the reference score $f$ is calculated from the aforementioned discrimination function $f(\cdot)$, $s$ is the learned anomaly score, and $\lambda_a$ is a weight hyperparameter to adjust the effect of score guidance for anomalies. $\epsilon_1$ and $\epsilon_2$ are thresholds to split the data space into obvious-normal and obvious-abnormal fields, as shown in Figure~\ref{fig:illustration}. As the training progresses, the anomaly scores of obvious-normal samples are expected to be zero and the scores of obvious-abnormal samples are expected to be large than value $a$. 
    However, the selection of the two thresholds is difficult in unsupervised settings. Considering that most samples in a dataset are normal, we apply only one threshold $\epsilon$ to divide the obvious-normal field and the suspected abnormal field. Then the regularization function $L_{\rm{SE}}$ is revised as follows:
    \begin{equation}\label{eq:SELoss}
        L_{\rm{SE}} (f, s) = 
        \begin{cases}
        |s-\mu_0|, & f < \epsilon \\
        \lambda_a \cdot \max(0, a-s), & f\ge \epsilon
        \end{cases}.
    \end{equation}
    We set a very small positive value $\mu_0$ approaching zero to be the target anomaly scores for the obvious-normal data, due to that the zero scores will force most weights of the scoring network to be $0$. 
    The threshold $\epsilon$ is a specific score value and difficult to choose, thus we select a percentile, denoted as $\epsilon_p$, of the reference scores to find the corresponding $\epsilon$. For example, we can set $\epsilon_p=70\%$ if we assume that 70\% of data are obvious-normal samples. Then the samples with scores less than $0.7\cdot f$ are regarded as obvious-normal ones and guided to obtain smaller scores. The other samples are considered as suspected abnormal ones and tend to get larger scores. In fact, $\epsilon_p$ does not require an accurate normal data ratio, which is unknown in the unsupervised scenario. Because it is an approximate proportion to define the obvious-normal samples and can be much smaller than the true normal data ratio, which has been confirmed in the experiments and reflected in Figure~\ref{fig:heatmap}. 
    We further discuss the hyperparameter selection for general datasets in Section~\ref{sec:param}.
    
    Note that some samples, especially in the transition field, may be judged as suspected abnormal data by the pre-defined metric, and their anomaly scores will be guided in the wrong direction in the early stage of training. However, there are also obvious-normal and obvious-abnormal samples that are effortless to identify. The score-guided model can leverage these obvious data to accelerate the convergence of the model and enhance the capability of the representation learner to extract more valuable information that more clearly reflects the abnormality of the data. With the gradually enhanced detection capability during the training process, the score-guided model can finally be expected to better distinguish these non-obvious data and assign them correct anomaly scores. 

    The score-guided regularization can be integrated with existing URL methods. Let $L_{\rm{OE}}$ be the loss function of an URL model, the total loss function $L$ of the integrated model with score-guided regularization is  
    \begin{equation}\label{eq:TotalLoss}
        L = L_{\rm{OE}} + \lambda_{\rm{SE}} \cdot L_{\rm{SE}},
    \end{equation}
    where $\rm{\lambda_{SE}}$ is a hyperparameter to adjust the effect of score-guided regularization. In other words, it affects the convergence speed of the $L_{\rm{SE}}$ relative to the $L_{\rm{OE}}$. $L_{\rm{OE}}$ ensures the optimization of the representation learner, while $L_{\rm{SE}}$ directly guides the training of the scoring network and indirectly promotes the representation learner to obtain differentiated representation. $\rm{\lambda_{SE}}$ is usually less than 1 to ensure faster convergence of the representation learner than the scoring network. These two loss functions participate in the training of the model together.
    
    The score-guided regularizer is easy to combine limited labels by using labels instead of reference scores for the labeled data. In this case, the regularizer does not rely entirely on labels to guide the anomaly scores like DevNet, due to the use of self-supervised signals.

\subsection{Score-guided Models}
\label{subs:sgm}
    
    As mentioned above, we can integrate the scoring network into unsupervised representation learning based methods. The score-guided methods can be collectively called SG-Models. Here, we first propose SG-AE, an instantiation of applying scoring network to one reconstruction-based method, and then introduce a general form of SG-Models. 
    
\begin{figure}[htb!]
        \begin{center}
        \includegraphics[width=0.4\textwidth]{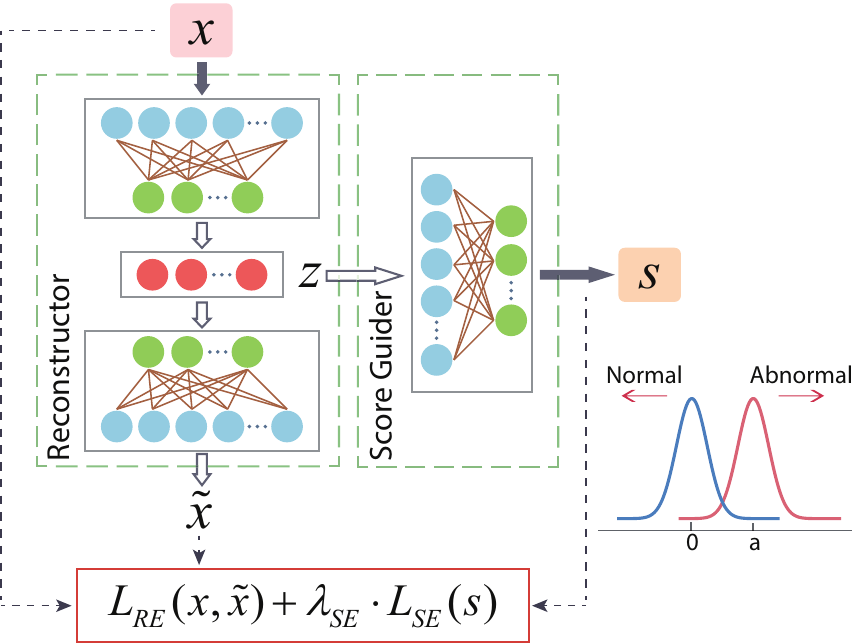}
        \end{center}
        \caption{The proposed score-guided autoencoder (SG-AE).}
        \label{fig:sgae}
    \end{figure}

    As illustrated in Figure~\ref{fig:sgae}, SG-AE consists of a reconstructor and a score guider. Given a data sample $\boldsymbol{x}$, the reconstructor first maps it to a representation $\boldsymbol{z}$ in a latent space and then generate an estimation $\boldsymbol{\tilde{x}}$. The score guider learns an anomaly score $\boldsymbol{s}$ using $\boldsymbol{z}$ as its input. 
    In practice, we adopt an autoencoder framework as our reconstructor and a fully connected network in the score guider. For tabular data, we also use fully connected networks in the autoencoder. It's noteworthy that this autoencoder network is easy to transfer to other data types. For example, we can utilize recurrent networks for sequence data and convolutional networks for image data~\cite{devnet,rsrae}.
    
    Let $\mathcal{E}$ and $\mathcal{D}$ denote the encoder and decoder in an autoencoder, respectively. We denote the score networking as $\mathcal{S}$, and thus have
    \begin{equation}\label{eq:forward}
        \boldsymbol{z} = \mathcal{E}(\boldsymbol{x}),\quad \boldsymbol{\tilde{x}} = \mathcal{D}(\boldsymbol{z}),\quad \boldsymbol{s} = \mathcal{S}(\boldsymbol{z}),
    \end{equation}
    where $\boldsymbol{x}$ is the input data; $\boldsymbol{\tilde{x}}$ is the reconstructed counterpart of $\boldsymbol{x}$; $\boldsymbol{z}$ is the latent representation; and $\boldsymbol{s}$ is the anomaly score. 
    
    The end-to-end network SG-AE can be represented as $\mathcal{M}$:
    \begin{equation}\label{eq:sgae}
        \mathcal{M}(\boldsymbol{x}) = \mathcal{S}(\mathcal{E}(\boldsymbol{x})). 
    \end{equation}
    We can obtain the parameters of $\mathcal{M}$ by minimizing a loss function, which is a combination of two parts: the reconstruction loss and the score-guidance loss. For the autoencoder, we use the $\mathcal{L}_2$ loss to assess the reconstruction, thus the $L_{\rm{OE}}$ can be represented as, 
    \begin{equation}\label{eq:AELoss}
        L_{\rm{OE}} (\boldsymbol{x}, \boldsymbol{\tilde{x}}) = \frac{1}{N} \sum_{i=1}^N  \big\| \boldsymbol{x}_i - \boldsymbol{\tilde{x}}_i \big \|_2.
    \end{equation}
    
    In the reconstruction-based settings, we also utilize the reconstruction error as a self-supervised signal, that is, we use $\mathcal{L}_2$ as the self-signal function $f(\cdot)$. Thus, we can rewrite the score-guided regularization in Eq.~(\ref{eq:SELoss}) as,
    \begin{equation}\label{eq:SELoss1}
    \resizebox{.85\linewidth}{!}{$
        L_{\rm{SE}} (\big\| \boldsymbol{x}_i - \boldsymbol{\tilde{x}}_i \big\|_2, s_i) = 
        \begin{cases}
        |s-\mu_0|, & \big\| \boldsymbol{x}_i - \boldsymbol{\tilde{x}}_i \big\|_2 < \epsilon \\
        \lambda_a \cdot \max(0, a-s), & \big\| \boldsymbol{x}_i - \boldsymbol{\tilde{x}}_i \big\|_2 \ge \epsilon
        \end{cases}
    $}.
    \end{equation}
    We finally define the overall loss function of SG-AE, an instantiation of Eq.~(\ref{eq:TotalLoss}), as the combination of the two loss items in Eq.~(\ref{eq:AELoss}) and~(\ref{eq:SELoss1}), that is,
    \begin{equation}\label{eq:Loss}
        L(\boldsymbol{x}, \boldsymbol{\tilde{x}}, s) = \frac{1}{N} \sum_{i=1}^N  \big\| \boldsymbol{x}_i - \boldsymbol{\tilde{x}}_i \big \|_2 + \lambda_{\rm{SE}} \cdot
        L_{\rm{SE}} (\big\| \boldsymbol{x}_i - \boldsymbol{\tilde{x}}_i \big\|_2, \boldsymbol{s_i}).
    \end{equation}
    
    Our goal is to minimize the loss function~(\ref{eq:Loss}). Lower reconstruction loss in Eq.~(\ref{eq:AELoss}) ensures better reconstructions $\boldsymbol{\tilde{x}}$ and representations $\boldsymbol{z}$ of normal data. As the score-guidance loss decreases, the anomaly scores of normal data approach zero and the scores of suspected abnormal data tend to $a$, thus the score disparities between normal and abnormal data continue to widen. A higher anomaly score of a sample indicates a higher possibility to be an anomaly. By observing the distribution of anomaly scores, one can distinguish between normal and abnormal samples.
    
    In addition to SG-AE with reconstruction-based assumption, the scoring network can also be applied to other URL methods, such as distance-based methods and distribution-based methods. Here, we discuss a general form of combining the scoring network with URL methods. Let $\mathcal{R}$ be a representation learner in an URL method. With the discrimination metric $f(\cdot)$, the general form of an unsupervised method is:
    \begin{equation}\label{eq:sgm1}
        \mathcal{M}_{URL}(\boldsymbol{x}) = f(\mathcal{R}(\boldsymbol{x})). 
    \end{equation}
    Combined with the scoring network $\mathcal{S}$, the general form of SG-Models is:
    \begin{equation}\label{eq:sgm2}
        \mathcal{M}_{SG}(\boldsymbol{x}) = \mathcal{S}(\mathcal{R}(\boldsymbol{x})). 
    \end{equation}
    The metric value from $f(\cdot)$ in SG-Models is only used as a self-supervised signal to assist model training and does not participate in the calculation of anomaly score. 
    By optimizing the total loss function, Eq.~(\ref{eq:TotalLoss}), the score-guided regularization can encourage SG-Models to learn and guide the anomaly scores, and also help the representation learner of the original model to learn better data representations. 
    To better understand the transferability, we extend the scoring network to four state-of-the-art models and evaluate the performance in the experiments.

\subsection{The SG-Models Algorithm}
    We illustrate the training process of SG-Models in Algorithm~\ref{alg:sgm}. 
    An SG-Model $\mathcal{M}_{SG}$ consists of an existing URL model $\mathcal{M}_{URL}$ and the scoring network $\mathcal{S}$. The $\mathcal{M}_{URL}$ has a representation learner $\mathcal{R}$, a pre-defined metric function $f(\cdot)$, and its original loss function $L_{\rm{OE}}$. The metric value calculated by $f(\cdot)$ is denoted as $f$.
    The parameters of $\mathcal{M}_{SG}$ are initialized randomly and optimized in the training iteration (Steps 2-12) with the minimization of the loss in Eq.~(\ref{eq:TotalLoss}). Specifically, the data representation $\boldsymbol{z}$ is learned by the representation learner $\mathcal{R}$, through $\boldsymbol{z} = \mathcal{R}(\boldsymbol{x})$ in Step 5. Then the scoring network learns the anomaly scores $\boldsymbol{s}$ through $\boldsymbol{s} = \mathcal{S}(\boldsymbol{z})$ in Step 6. 
    In Step 7, we calculate $L_{\rm{OE}}$ and the self-supervised signal $f$ from $\mathcal{M}_{URL}$. Then Step 8 calculates $L_{\rm{SE}}$ with the loss function~(\ref{eq:SELoss}). After that, we get the total loss in Step 9, and Step 10
     takes the back-propagation algorithm and updates the network parameters. With the guidance from the loss function~(\ref{eq:TotalLoss}), data representation and anomaly score distribution are optimized during training, and finally, we obtain a well-trained SG-Model.
    
    In terms of testing, the trained SG-Model $\mathcal{M}_{SG}$ directly calculates the anomaly scores $\boldsymbol{s}$ given samples $\boldsymbol{x}$, that is, $\boldsymbol{s} = \mathcal{M}_{SG}(\boldsymbol{x}) = \mathcal{S}(\mathcal{R}(\boldsymbol{x}))$. The pre-defined metric function $f(\cdot)$ is not used in testing. The anomaly scores reflect the possibility of being anomalies. A sample with a higher anomaly score means that it has a greater possibility of being abnormal.

\begin{algorithm}
    \caption{Training an SG-Model}
    \begin{algorithmic}[1]
    \label{alg:sgm}
        \renewcommand{\algorithmicrequire}{\textbf{Input:}}
        \renewcommand{\algorithmicensure}{\textbf{Output:}} 
        \REQUIRE Training set $\{\boldsymbol{x}_i\}_N$, $\boldsymbol{x}_i\in\mathbb{R}^D$
        \ENSURE  An SG-Model $\mathcal{M}_{SG}$:An optimized anomaly detection network
        \STATE Randomly initialize the network parameters of $\mathcal{M}_{URL}$, $\mathcal{S}$
        \FOR{each epoch}
            \STATE {Divide input data into batches}
            \FOR {each batch}
                \STATE $\boldsymbol{z} \leftarrow \boldsymbol{z} = \mathcal{R}(\boldsymbol{x})$ 
                \STATE $\boldsymbol{s} \leftarrow \boldsymbol{s} = \mathcal{S}(\boldsymbol{z})$
                \STATE $L_{\rm{OE}}, f \leftarrow \mathcal{M}_{URL}$
                \STATE $L_{\rm{SE}} \leftarrow L_{\rm{SE}}(f, \boldsymbol{s})$
                \STATE $L$ $\leftarrow L = L_{\rm{OE}} + L_{\rm{SE}}$
                \STATE Backpropagate and update the parameters of $\mathcal{M}_{URL}$, $\mathcal{S}$ 
        \ENDFOR
        \ENDFOR
        \RETURN An optimized SG-Model
     \end{algorithmic}

\end{algorithm}

\section{Experiments}
    In this section, we empirically evaluate the effectiveness and robustness of the SG-Models on both synthetic and real-world datasets.

\subsection{Simulation Experiments}

    The score-guided regularization is designed to intensify the differentiation between normal and abnormal samples during the training process.
    To verify that, we conduct simulation experiments to observe the change in anomaly scores during the model training. We compare our SG-AE model with AutoEncoder (AE)~\cite{ae} and RDP~\cite{rdp} in three artificial datasets.
    
\begin{figure*}[bt!]
        \begin{center}
        \includegraphics[width=0.85\textwidth]{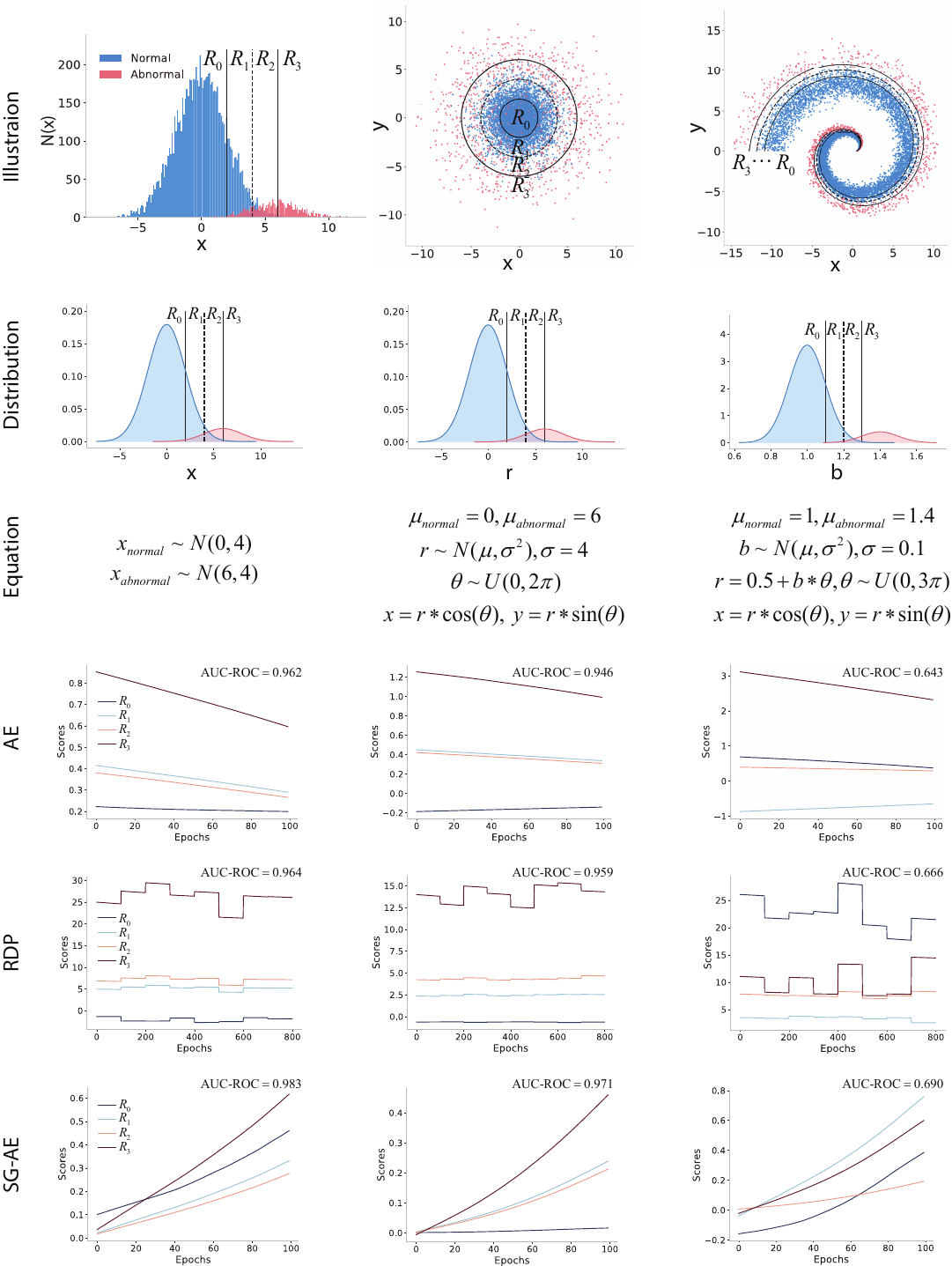}
        \end{center}
        \caption{Simulation experiments in three synthetic datasets. $({\rm{R}}_0{\sim}{\rm{R}}_3)$ represent fields with different ratio of anomaly data. For SG-AE, the score difference between normal and abnormal data in each field gradually expands with the training process.}
        \label{fig:simulation}
    \end{figure*}

    \textbf{Data generation}. As illustrated in Figure~\ref{fig:simulation}, we generate a one-dimensional dataset and two two-dimensional datasets. The normal data are shown in blue while the abnormal data are in red.  
    We utilize different equations to generate datasets of different complexity. For the one-dimensional scenario, we sample data points from Gaussian distribution $\mathcal{N}(\mu,\sigma^{2})$ in geometric coordinates. To generate samples in two-dimensional scenarios, we control two variables in polar coordinates and then transform them into geometric coordinates. One variable follows a Gaussian distribution and the other follows a uniform distribution. The normal and abnormal data are sampled with different mean $\mu$ and the same variance $\sigma^{2}$. The detailed equations are shown in Figure~\ref{fig:simulation}. 
    
    \textbf{Settings}. 
    The training and testing sets are independently generated, each with 10,000 data samples. The ratio of abnormal to normal data is one to nine. In the testing sets, we use three thresholds to divide each dataset into four parts $({\rm{R}}_0{\sim}{\rm{R}}_3)$, as shown in Figure~\ref{fig:simulation}. 
    For a normal distribution, the more off-center a sample is, the less likely it is to appear and the more likely it is to be an anomaly. Therefore, we could use the 3-$\sigma$ rule to select the thresholds, that is $\{\mu+\sigma, \mu+2\sigma, \mu+3\sigma \}$
    Most of the data in ${\rm{R}}_0$ are normal, namely \textit{obvious}-\textit{normal}. Similarly, ${\rm{R}}_3$ is regarded as \textit{obvious}-\textit{abnormal}; ${\rm{R}}_1$ and ${\rm{R}}_2$ are two transition fields.
    In each field, we calculate the difference between the average anomaly scores of abnormal data and normal data. That is,
    \begin{equation}
        S_i = \frac{1}{N_{ai}} \sum_{j=1}^{N_{ai}} s_j - \frac{1}{N_{ni}} \sum_{k=1}^{N_{ni}} s_k, i \in [0, 1, 2, 3],
    \end{equation}
    where $S_i$ is the score difference in ${\rm{R}}_i$, $N_{ai}$ and $N_{ni}$ are the numbers of abnormal and normal samples in ${\rm{R}}_i$, $s_j$ is the anomaly score of a sample.
    Note that the scores of these models are calculated in different ways, thus we mainly compare the trends of the score differences during training, which reflect the capability of the model to learn the disparities between normal and abnormal data.
    We also use the Area Under Receiver Operating Characteristic Curve (AUC-ROC) to evaluate the model performances. In terms of model settings, AE and SG-AE share the same Adam optimizer, learning rate, and fully connected networks. The layer sizes are set as 20 for the one-dimensional dataset and (64, 20) for the two-dimensional datasets. The scoring network in SG-AE is also a fully connected network with layer size (20, 1). The RDP keeps the originally proposed settings. The batch size and the training epoch are 1024 and 100, respectively.
    
    \textbf{Results analysis}. As shown in Figure~\ref{fig:simulation}, for AE, the score differences decrease during training, suggesting that AE attempts to reconstruct the abnormal data. The continuous decrease of score differences $S_1$ and $S_2$ means that AE starts to fail in distinguishing the normal and abnormal data in the transition field. For RDP, it uses a boosting process to filter anomalies and the iteration size is set as 8, thus the curves of score difference are jagged and the total epochs are 800. The changes of $S_1$ and $S_2$ are not obvious. For SG-AE, the score difference in each field is increasing during training, suggesting that the model reacts differently to normal and abnormal data. Specifically, the abnormal data are guided to obtain a larger anomaly score, while the normal data to a smaller anomaly score. More important, as expected, the increase of $S_1$ and $S_2$ indicates the improvement of the detection ability of the model in the transition field. Moreover, the SG-AE achieves the best AUC-ROC performances in the three datasets. The results demonstrate the effectiveness of the scoring network.

\subsection{Evaluation Experiments} \label{subsec:eval}
\subsubsection{Datasets Description}

    As shown in Table~\ref{tab:data}, we use seven publicly available tabular datasets in our experiments. Two datasets are in the healthcare domain, e.g., diagnosis of breast cancer (\textit{bcsc})~\cite{bcsc} and diabetic detection (\textit{diabetic})~\cite{diabetic}. Another two datasets are related to the attack in cybersecurity, e.g., \textit{intrusion}~\cite{uci} and \textit{attack}~\cite{attack}; and  \textit{Market} includes data of potential subscribers in bank marketing~\cite{misc_bank_marketing_222}. 
    \textit{Creditcard} is a credit-card fraud dataset. \textit{Donor} is a valuable project selection dataset. Details of these two datasets can be found in~\cite{devnet}. The notations in Table~\ref{tab:data} are defined as follows: $\rm{N_{origin}}$ is the original size of the dataset. $\rm{D_{num}}$ and $\rm{D_{cat}}$ are dimensions of numerical and categorical features. $\rm{R_{dul}}$ and $\rm{R_{miss}}$ are the rates of duplicate and missing data. $\rm{N}$ and $\rm{D}$ are the size and dimension after preprocessing. $\rm{N_{noise}}$ is the number of injected noise and $\rm{R}_{anomaly}$ is the rate of anomalies.

\begin{table}[hbt!]
\caption{Statistical information of datasets.} 
\label{tab:data}
\centering
\scalebox{0.73}{
    \setlength\tabcolsep{3pt}
    \begin{tabular}{l|rrrrr|rrrr}
    \toprule
        & 
        \multicolumn{5}{c|}{\textbf{Before Preprocessing}} & 
        \multicolumn{4}{c}{\textbf{After Preprocessing}} \\
    \midrule
        \textbf{Data}       & \boldmath$\rm{N_{origin}}$ & \boldmath$\rm{D_{num}}$ & \boldmath$\rm{D_{cat}}$ & \boldmath$\rm{R_{dul}}$(\%) & \boldmath$\rm{R_{miss}}$(\%) & \boldmath$\rm{N}$     & \boldmath$\rm{D}$  & \boldmath$\rm{N_{noise}}$ & \boldmath$\rm{R}_{anomaly}$(\%) \\
    \midrule
        Attack     & 257673  & 39   & 3    & 0.00      & 0.00        & 75965  & 191 & 342    & 25.24    \\
        Bcsc       & 462563  & 14   & 0    & 17.87     & 0.00        & 382122 & 14  & 2242   & 2.62     \\
        Creditcard & 284807  & 30   & 0    & 0.38      & 0.00        & 285441 & 30  & 1716   & 0.17     \\
        Diabetic   & 101766  & 12   & 31   & 0.00      & 3.65        & 98575  & 115 & 522    & 12.10    \\
        Donor      & 664098  & 6    & 16   & 0.04      & 12.02       & 587383 & 79  & 3317   & 6.28     \\
        Intrusion  & 805050  & 38   & 3    & 73.22     & 0.00        & 216420 & 119 & 820    & 37.23    \\
        Market     & 45211   & 7    & 9    & 0.00      & 0.00        & 45451  & 51  & 241    & 11.70    \\
    \bottomrule
    \end{tabular}
}
\end{table}

    \textbf{Data Preprocessing}. We first preprocess the datasets by removing duplicated samples or samples with missing features. For example, 73.22\% of duplicates are found on \textit{intrusion}, and 12.02\% of missing data are found on \textit{donor}. 
    Next, we encode the categorical features with one-hot encoding and standardize each numerical feature. In addition, following~\cite{devnet} we inject noises into the training data to improve the robustness. Specifically, we randomly select 1\% of normal data and swap 5\% features of these data. Finally, the training, validation, and testing sets are randomly divided into 6/2/2. 
    Note that some datasets have a part of features that leak anomaly labels. To keep consistency in real applications, we drop one feature in \textit{attack} and five features in \textit{donor} to avoid using this supervised-like information.

\subsubsection{Benchmarks and Settings}

    We compare SG-AE with eight benchmarks: 
    two statistical learning methods, iForest~\cite{iforest} and ECOD~\cite{li2022ecod}, and six deep learning methods, RDA~\cite{rda}, DAGMM~\cite{dagmm}, RDP~\cite{rdp}, RSRAE~\cite{rsrae}, GOAD~\cite{goad}, and ICLAD~\cite{shenkar2022anomaly}. We use the originally proposed model structures for iForest, DAGMM, RDP, GOAD, ECOD, and ICLAD. As for RDA, RSRAE, and SG-AE, we use the same fully connected layer settings in the autoencoder framework to maintain consistency.

    In terms of hyperparameter settings, these two statistical learning methods, iForest and ECOD, do not require hyperparameters. 
    The deep learning models are sensitive to their hyperparameters in different datasets. To fairly compare the upper bound of performance, we conduct hyperparameter searching to obtain optimal results, starting from the recommended hyperparameters. 
    The mainly tuned hyperparameters of deep baselines are described as follows:
    \begin{itemize}
        \item RDA isolates suspected anomalies and fits cleaned samples. It has a $\lambda$ to adjust the proportion of isolated samples and a hyperparameter to control the step interval to update the isolation.  
        \item DAGMM uses Gaussian Mixture Model to fit data representation. It has two $\lambda$ to adjust loss weights and a hyperparameter to decide the component number of Gaussian Mixture Model.
        \item RSRAE utilizes a robust subspace recovery (RSR) layer to remove anomalies in the hidden space. It is tuned with two loss weights $\lambda$ and the intrinsic dimension of latent space. 
        \item RDP adopts a training strategy of multiple iterations, filtering suspected anomalies and retraining the model at each iteration. It is tuned with a loss weight $\lambda$, the iteration number, and the latent dimension.
        \item GOAD learns a spherical distribution for data representation. It is tuned with a loss weight $\lambda$, and two dimension-related hyperparameters.
        \item ICLAD maximizes the mutual information between one sample and its masked counterpart. It is tuned with three hyperparameters, including the kernel size, the hidden dimension, and the temperature constant in its loss.
    \end{itemize}
    
    SG-AE is tuned with two $\lambda$ and $\epsilon_p$, while $a$ is set to 6. We set batch size as $1024$, the training epochs as $100$, and the Adam optimizer with a learning rate of $0.0001$. We select the well-trained models in the validation sets. The PyTorch implementation code is publicly released at https://github.com/urbanmobility/SGM and we also implement the code with MindSpore released at https://github.com/urbanmobility/SGM\_mindspore.

\subsubsection{Performance Evaluation}

    We use the Area Under Receiver Operating Characteristic Curve (AUC-ROC) and the Area Under Precision-Recall Curve (AUC-PR) to evaluate the performance of reference methods. Higher AUC values suggest better performance. The results are averaged over 10 independent runs, in which the datasets are regenerated randomly, to report stable performance and are shown in Table~\ref{tab:performance}. The best performance in each dataset is highlighted in bold text and the second-best one is underlined.

    Among these baselines, we can see RDP and ECOD achieve the best and second-best performances in terms of AUC-ROC. DAGMM, GOAD, and ICLAD attempt to fit all training data without handling suspected anomalies, thus their performances are easily degraded by noise and data pollution. On the contrary, RDP gradually filters out the suspected anomalies and fits the rest of the data during training, while ECOD uses empirical cumulative distribution functions to estimate the anomaly probability of each sample. These strategies to deal with data contamination contribute to the relatively better performances of RDP and ECOD. 
    While RDA and RSRAE also use filtering strategies, they are not as effective as RDP due to their less reliable filtering techniques. 
    In contrast, SG-AE utilizes an active strategy, called the score-guidance strategy, to handle contaminated samples. Unlike these passive strategies used in baselines, our approach actively utilizes the samples with high abnormal probabilities to guide model optimization, thereby enhancing both the capability of the representation learner and the score disparity. This is the main reason why SG-AE outperforms these baselines in all datasets. 

    Compared with the results in~\cite{rdp}, the performance difference is mainly caused by the construction of the datasets. For example, we obtained 22 features from the data sources in \textit{donor} but~\cite{rdp} takes only 10 features without detailed prepossessing steps. However, we can have a convincing comparison result in \textit{creditcard}, a standard dataset preprocessed in the original data source. On this dataset, iForest and RDP perform similarly in our work and~\cite{rdp}.

\begin{table*}[hbt]
\caption{AUC-ROC and AUC-PR (mean$\pm$std) of SG-AE and the reference models.}
\label{tab:performance}
\centering
\scalebox{0.86}{
    \setlength\tabcolsep{4pt}
    \begin{tabular}{c|l|ccccccccc}
    \toprule
        \textbf{Metrics} & \textbf{Datasets} &
        \textbf{iForest} &  \textbf{DAGMM} & \textbf{RDP} & \textbf{RDA} &
            \textbf{RSRAE} & \textbf{GOAD} & \textbf{ECOD} & \textbf{ICLAD} & \textbf{SG-AE} \\
    \midrule    
        \multirow{7}{*}{\textbf{AUC-ROC}} &
        Attack     & 0.644$\pm$0.028 & 0.762$\pm$0.012 & 0.769$\pm$0.016 & 0.615$\pm$0.033 & 0.681$\pm$0.035 
                    & 0.730$\pm$0.003 & \underline{0.770$\pm$0.000} & 0.746$\pm$0.018
                    & \textbf{0.833$\pm$0.009} \\
        & Bcsc       & 0.754$\pm$0.033 & 0.672$\pm$0.023 & \underline{0.894$\pm$0.003} & 0.872$\pm$0.004 & 0.840$\pm$0.023 
                    & 0.882$\pm$0.004 & 0.749$\pm$0.000 & 0.800$\pm$0.015
                    & \textbf{0.912$\pm$0.008} \\
        & Creditcard & 0.952$\pm$0.003 & 0.833$\pm$0.057 & \underline{0.953$\pm$0.005} & 0.920$\pm$0.006 & 0.942$\pm$0.017 
                     & 0.939$\pm$0.013 & 0.934$\pm$0.000 & 0.933$\pm$0.008
                    & \textbf{0.964$\pm$0.004} \\
        & Diabetic   & \underline{0.546$\pm$0.008} & 0.514$\pm$0.020 & 0.541$\pm$0.041 & 0.538$\pm$0.005 & 0.477$\pm$0.004 
                    & 0.524$\pm$0.013 & 0.539$\pm$0.000 & 0.522$\pm$0.005
                    & \textbf{0.548$\pm$0.007} \\
        & Donor      & 0.509$\pm$0.005 & 0.519$\pm$0.011 & 0.518$\pm$0.021 & 0.513$\pm$0.001 & 0.484$\pm$0.009 
                    & 0.495$\pm$0.007 & 0.499$\pm$0.000 & \underline{0.537$\pm$0.009}
                    & \textbf{0.541$\pm$0.004} \\
        & Intrusion  & 0.815$\pm$0.012 & 0.860$\pm$0.046 & 0.863$\pm$0.022 & 0.614$\pm$0.027 & 0.741$\pm$0.022 
                    & 0.898$\pm$0.008 & \underline{0.903$\pm$0.000} & 0.728$\pm$0.010
                    & \textbf{0.906$\pm$0.012} \\
        & Market     & 0.652$\pm$0.016 & 0.640$\pm$0.013 & 0.685$\pm$0.02 & 0.674$\pm$0.006 & 0.536$\pm$0.013 
                     & \underline{0.709$\pm$0.018} & 0.676$\pm$0.000 & 0.704$\pm$0.016
                    & \textbf{0.750$\pm$0.010} \\
    \midrule
        \multirow{7}{*}{\textbf{AUC-PR}} &
        Attack      & 0.343$\pm$0.027 & 0.457$\pm$0.010 & 0.442$\pm$0.020 & 0.321$\pm$0.013 & \underline{0.533$\pm$0.046} 
                        & 0.404$\pm$0.003 & 0.457$\pm$0.000 & 0.165$\pm$0.005
                        & \textbf{0.596$\pm$0.014}  \\ 
        & Bcsc        & 0.080$\pm$0.026 & 0.053$\pm$0.011 & 0.605$\pm$0.024 & \underline{0.613$\pm$0.003} & 0.137$\pm$0.048 
                        & 0.603$\pm$0.014 & 0.075$\pm$0.000 & 0.016$\pm$0.000
                        & \textbf{0.816$\pm$0.004}  \\
        & Creditcard  & 0.184$\pm$0.031 & 0.100$\pm$0.185 & 0.187$\pm$0.032 & \underline{0.220$\pm$0.003} & 0.158$\pm$0.054 
                        & 0.115$\pm$0.023 & 0.161$\pm$0.000 & 0.001$\pm$0.000
                        & \textbf{0.291$\pm$0.042}  \\
        & Diabetic    & \underline{0.136$\pm$0.003} & 0.128$\pm$0.010 & 0.135$\pm$0.017 & 0.132$\pm$0.003 & 0.117$\pm$0.001 
                        & 0.126$\pm$0.005 & 0.132$\pm$0.000 & 0.113$\pm$0.002
                        & \textbf{0.142$\pm$0.004}  \\
        & Donor       & 0.066$\pm$0.001 & \underline{0.069$\pm$0.002} & 0.067$\pm$0.005 & 0.066$\pm$0.000 & 0.060$\pm$0.001 
                        & 0.063$\pm$0.002 & 0.064$\pm$0.000 & 0.058$\pm$0.002
                        & \textbf{0.075$\pm$0.002}  \\
        & Intrusion   & 0.619$\pm$0.017 & \underline{0.801$\pm$0.034} & 0.666$\pm$0.027 & 0.478$\pm$0.012 & 0.747$\pm$0.042
                        & 0.713$\pm$0.018 & 0.762$\pm$0.000 & 0.264$\pm$0.003
                        & \textbf{0.904$\pm$0.035}  \\
        & Market      & 0.189$\pm$0.011 & 0.211$\pm$0.007 & \underline{0.227$\pm$0.022} & 0.210$\pm$0.006 & 0.129$\pm$0.005 
                        & 0.224$\pm$0.021 & 0.215$\pm$0.000 & 0.077$\pm$0.002
                        & \textbf{0.270$\pm$0.020}  \\
    \bottomrule
    \end{tabular}
}
\end{table*}    

\begin{figure}[hbt!]
\begin{center}
\includegraphics[width=0.48\textwidth]{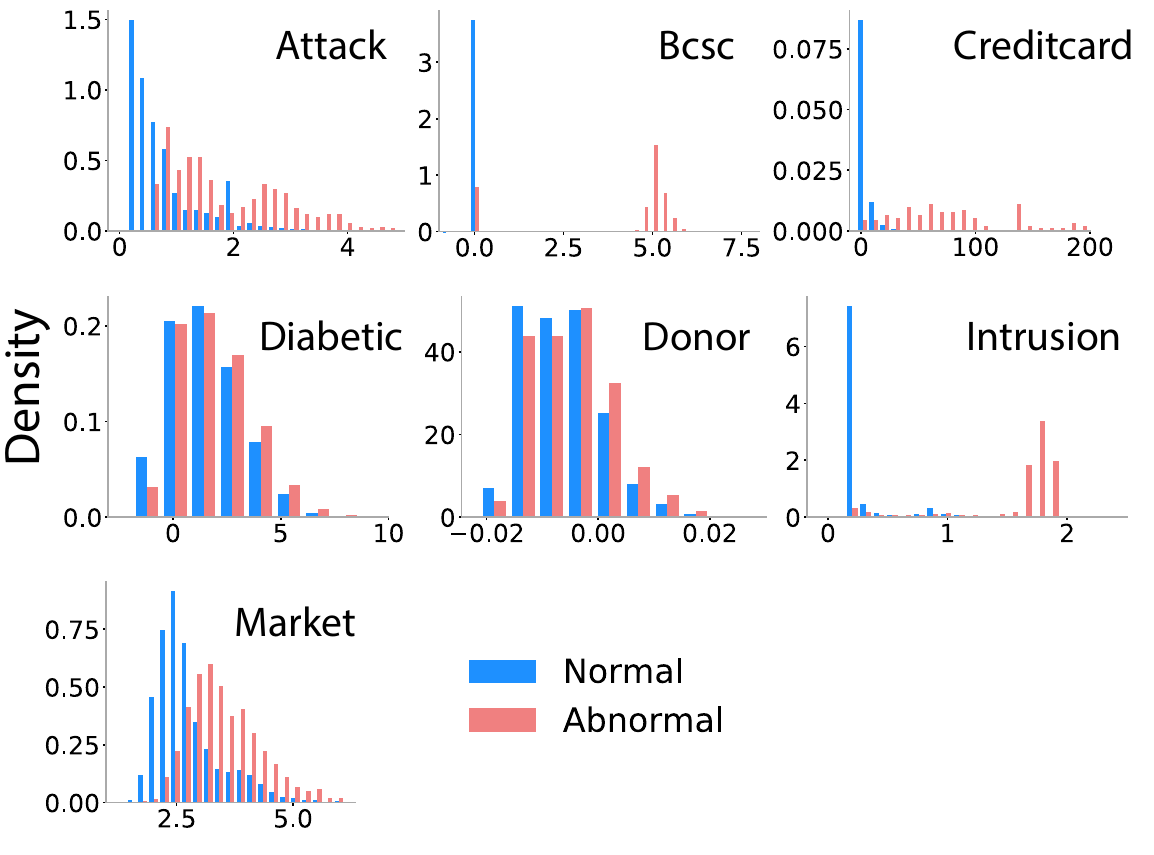}
\end{center}
\caption{The anomaly score distribution of actual normal and abnormal data on the seven datasets. As expected, the SG-AE learns relative large anomaly scores for the abnormal data (red), but small scores for the normal data (blue).}
\label{fig:hist}
\end{figure}

    Figure~\ref{fig:hist} presents the score distribution of SG-AE. We can see that the anomaly score distributions of normal data (blue bars) and abnormal data (red bars) are notably separated on most datasets. 
    For numerical comparison between the typical methods, we use the KS test to reflect the distribution distance between normal and abnormal scores. KS test is a statistical hypothesis test that checks whether two samples differ in distributions~\cite{dos2016fast}. A higher KS index indicates a larger score difference. As shown in Table~\ref{tab:KSindex}, the average values of the KS of SG-AE on the seven datasets are $0.539$, surpassing the $0.355$ of AE, $0.372$ of AE, $0.336$ of RDA, $0.415$ of RDP, $0.227$ of GOAD, $0.398$ of ECOD, and $0.362$ of ICLAD. This suggests that our scoring network successfully enhanced the disparity between the normal and abnormal data. A larger KS index correlates with a better performance shown in Table~\ref{tab:performance}, especially on the \textit{bcsc}, \textit{creditcard}, \textit{intrusion} datasets. Taken together, these results demonstrate the effectiveness of the score guidance design. 
    
\begin{table}[htb!]
\caption{KS index between the anomaly score distributions of the normal and abnormal data in various models.} 
\label{tab:KSindex}
\centering
\scalebox{0.85}{
    \setlength\tabcolsep{3pt}
    \begin{tabular}{l|ccccccc}
    \toprule
        \textbf{KS index} & \textbf{Attack} & \textbf{Bcsc} & \textbf{Creditcard} & \textbf{Diabetic} & \textbf{Donor} & \textbf{Intrusion} & \textbf{Market} \\ 
    \midrule
        \textbf{iForest} & 0.207  & 0.456   & 0.805  & 0.076    & 0.035  & 0.717 & 0.188 \\
        \textbf{AE} & 0.224  & 0.089   & 0.857  & 0.082    & 0.086  & 0.843 & 0.422 \\
        \textbf{RDA} & 0.211  & 0.760   & 0.782  & 0.064    & 0.024  & 0.214 & 0.265 \\
        \textbf{RDP} & 0.346  & 0.771  & 0.870  & 0.103    & 0.056  & 0.450 & 0.340 \\
        \textbf{GOAD} & 0.218 & 0.302 & 0.442 & 0.039 & 0.065 & 0.260 & 0.263 \\
        \textbf{ECOD} & 0.447 & 0.446 & 0.781 & 0.064 & 0.014 & 0.773 & 0.262 \\
        \textbf{ICLAD} & 0.420 & 0.499 & 0.785 & 0.037 & 0.059 & 0.469 & 0.266 \\
        \textbf{SG-AE} & \textbf{0.570}  & \textbf{0.791}  & \textbf{0.882}  & \textbf{0.114}    & \textbf{0.089}  & \textbf{0.864} & \textbf{0.509} \\
    \bottomrule
    \end{tabular}
}
\end{table}

\begin{figure*}[htbp]
\begin{center}
\includegraphics[width=0.85\textwidth]{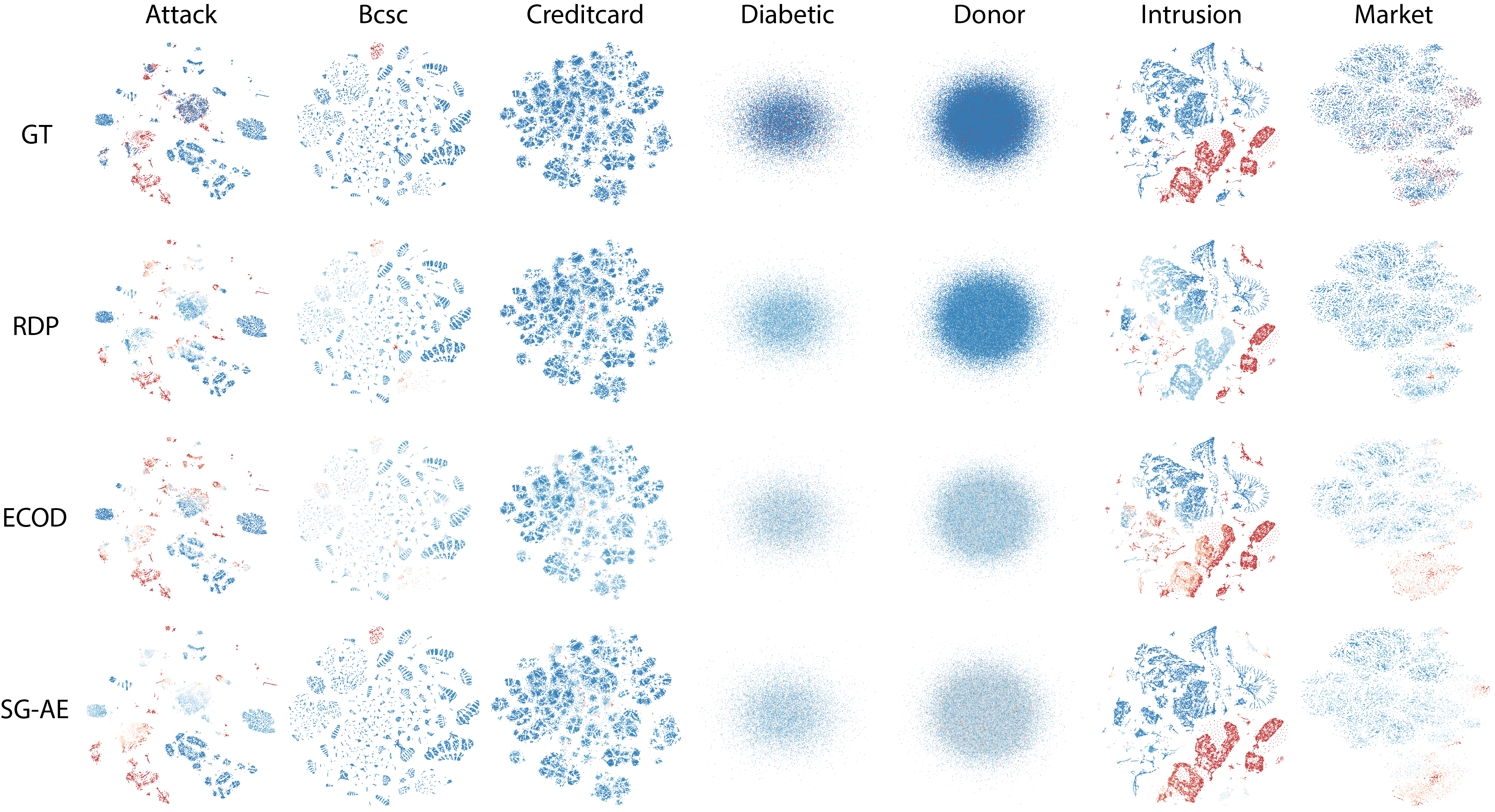}
\end{center}
\caption{The t-SNE visualization analysis. The three rows refer to the ground truth (GT) and results of RDP, ECOD}, and SG-AE, respectively. Red represents abnormal, while blue represents normal. The darker color indicates a larger probability of being normal or abnormal.
\label{fig:sne}
\end{figure*}

    To intuitively understand the anomaly detection results, we utilize t-SNE for dimensionality reduction and visualize the seven datasets in 2D space, as shown in Figure~\ref{fig:sne}. Data samples are colored with their true labels or anomaly scores generated by our SG-AE, RDP and ECOD. RDP and ECOD perform the best two among the baselines. The light color indicates these samples can not be easily distinguished as normal or abnormal data. One can select different thresholds to identify the abnormal samples, according to the datasets and needs. In Figure~\ref{fig:sne}, we observe that t-SNE can well segregate the anomalies in the 2D space on half of the datasets, and SG-AE presents more similar segregation to the ground truth than RDP and ECOD, especially on, \textit{attack}, \textit{bcsc} and \textit{intrusion}. These results are consistent with the results in Table~\ref{tab:performance}, reflecting the anomaly detection capabilities of SG-AE. We also notice that the anomalies are not segregated well on \textit{diabetic} and \textit{donor}, indicating the challenge of anomaly detection.    

\subsubsection{Comparison with Variants}

    In SG-AE, we introduce the score-guidance strategy to assign anomaly scores with the reconstruction error as a self-supervised signal. 
    To examine the effect of the learnable anomaly score, we remove the scoring network and use the reconstruction error as the anomaly score, keeping the regularizer in Eq.~(\ref{eq:Loss}). That means regularizer is used to guide the reconstruction error. This variant of SG-AE is named SG-AE$\rm{_{recon}}$. We also test AE, the original autoencoder without the score-guided loss and the scoring network.

    We next devise the other two variants, SG-AE$\rm{_{normal}}$ and SG-AE$\rm{_{lognormal}}$, to examine the score distribution assumption. The anomaly scores appear to have a normal or log-normal distribution in the statistical histogram displayed in Figure~\ref{fig:hist}. Inspired by this, we assume the anomaly scores follow a normal or lognormal distribution in our SG-AE. The outputs of the scoring network are altered to the mean and standard deviation of scores, denoted by $\mu$ and $\sigma$. We utilize KL divergence to measure the distribution difference, thus the score guided regularization (\ref{eq:SELoss1}) can be written as follows.
    \begin{equation}\label{eq:SELossVariants}
    \resizebox{.95\linewidth}{!}{$
        L_{\rm{SE}} (\boldsymbol{x}, \boldsymbol{\tilde{x}}, \mu, \sigma) =
        \begin{cases}
        \rm{KL} \big(   D(\mu,\sigma), D(0,1)  \big), & \big\| \boldsymbol{x} - \boldsymbol{\tilde{x}} \big\|_2 < \epsilon \\
        \lambda_a \cdot \max(0, a-\mu), & \big\| \boldsymbol{x} - \boldsymbol{\tilde{x}} \big\|_2 \ge \epsilon
        \end{cases}.
    $}
    \end{equation}

    Table~\ref{tab:variants} compares the performance of the original SG-AE and its variants. The original SG-AE performs best on five datasets. SG-AE$\rm{_{normal}}$ and SG-AE$\rm{_{lognormal}}$ have similar performances, which are slightly worse than SG-AE$\rm{_{original}}$ in most cases. SG-AE$\rm{_{recon}}$ achieves the best performances on \textit{diabetic} and \textit{donor}. However, the performance of SG-AE$\rm{_{recon}}$ has large variances across datasets, which is caused by the conflict of these two loss functions. The reconstruction loss tends to reduce the reconstruction error, but the score guidance loss attempt to enlarge the reconstruction error for anomalies. This conflict brings challenges to the update of network states. The results of SG-AE$\rm{_{original}}$, SG-AE$\rm{_{normal}}$, SG-AE$\rm{_{lognormal}}$ are significantly better than AE, which demonstrates the effectiveness of the scoring network.

\begin{table*}[htb]
\caption{AUC-ROC and AUC-PR (mean±std) of SG-AE and its variants.}
\label{tab:variants}
\centering
\scalebox{0.76}{
    \setlength\tabcolsep{5pt}
    \begin{tabular}{l|ccccc|ccccc}
    \toprule
        \multicolumn{6}{c|}{\textbf{AUC-ROC}} & 
        \multicolumn{5}{c}{\textbf{AUC-PR}} \\
    \midrule
        \textbf{Datasets} &
         \textbf{SG-AE}\boldmath$\rm{_{original}}$ &  \textbf{SG-AE}\boldmath$\rm{_{normal}}$ & \textbf{SG-AE}\boldmath$\rm{_{lognormal}}$ &  \textbf{SG-AE}\boldmath$\rm{_{recon}}$ & \textbf{AE} 
         & \textbf{SG-AE}\boldmath$\rm{_{original}}$  &  \textbf{SG-AE}\boldmath$\rm{_{normal}}$ &  \textbf{SG-AE}\boldmath$\rm{_{lognormal}}$ &  \textbf{SG-AE}\boldmath$\rm{_{recon}}$ & \textbf{AE}
        \\
    \midrule    
        \textbf{Attack}     & \textbf{0.833$\pm$0.009} & 0.772$\pm$0.06 & 0.745$\pm$0.09 & 0.639$\pm$0.112 & 0.561$\pm$0.037 &     \textbf{0.596$\pm$0.024} & 0.436$\pm$0.0724 & 0.412$\pm$0.079 & 0.436$\pm$0.106 & 	0.303$\pm$0.016 \\
        \textbf{Bcsc}       & \textbf{0.912$\pm$0.008} & 0.897$\pm$0.006 & 0.898$\pm$0.006 & 0.625$\pm$0.288 & 0.878$\pm$0.010 &     \textbf{0.816$\pm$0.004} & 0.781$\pm$0.050 & 0.751$\pm$0.078 & 0.229$\pm$0.281 & 	0.617$\pm$0.004  \\
        \textbf{Creditcard} & \textbf{0.964$\pm$0.004} & 0.940$\pm$0.017 & 0.928$\pm$0.027 & 0.716$\pm$0.158 & 0.903$\pm$0.013 &     \textbf{0.291$\pm$0.072} & 0.137$\pm$0.060 & 0.128$\pm$0.046 & 0.084$\pm$0.107 &	0.210$\pm$0.004  \\
        \textbf{Diabetic}   & 0.548$\pm$0.007 & 0.540$\pm$0.015 & 0.541$\pm$0.013 & \textbf{0.559$\pm$0.077} & 0.535$\pm$0.007 &     0.142$\pm$0.004 & 0.133$\pm$0.009 & 0.138$\pm$0.005 & \textbf{0.156$\pm$0.043} & 	0.131$\pm$0.003 \\
        \textbf{Donor}      & 0.541$\pm$0.004 & 0.532$\pm$0.006 & 0.536$\pm$0.006 & \textbf{0.559$\pm$0.063} & 0.486$\pm$0.045 &     0.075$\pm$0.002 & 0.072$\pm$0.002 & 0.073$\pm$0.002 & \textbf{0.076$\pm$0.012} & 	0.063$\pm$0.007  \\
        \textbf{Intrusion}  & \textbf{0.906$\pm$0.016} & 0.712$\pm$0.077 & 0.695$\pm$0.037 & 0.748$\pm$0.312 & 0.568$\pm$0.036 &      \textbf{0.904$\pm$0.035} & 0.546$\pm$0.055 & 0.541$\pm$0.024 & 0.730$\pm$0.242 & 	0.461$\pm$0.017  \\
        \textbf{Market}     & \textbf{0.750$\pm$0.025} & 0.647$\pm0.018$ & 0.644$\pm$0.013 & 0.600$\pm$0.076 & 0.677$\pm$0.008 &      \textbf{0.270$\pm$0.030} & 0.191$\pm$0.009 & 0.187$\pm$0.006 & 0.191$\pm$0.065 & 	0.215$\pm$0.007 \\
    \bottomrule
    \end{tabular}
}
\end{table*}

\subsubsection{Sensitivity of Parameters}
\label{sec:sens}

    In this part, we examine the sensitivity of SG-AE to different parameters. For convenience, we divide the four parameters into two groups, $\epsilon_p$ and $a$ in group one, $\rm\lambda_{SE}$ and $\lambda_{a}$ in group two. We first test the group one by fixing $\rm\lambda_{SE}=0.01$ and $\lambda_{a}=18$, then test the group two with fixed $\epsilon_p$ and $a$.
    
    \begin{figure*}[tb!]
        \begin{center}
        \includegraphics[width=0.85\textwidth]{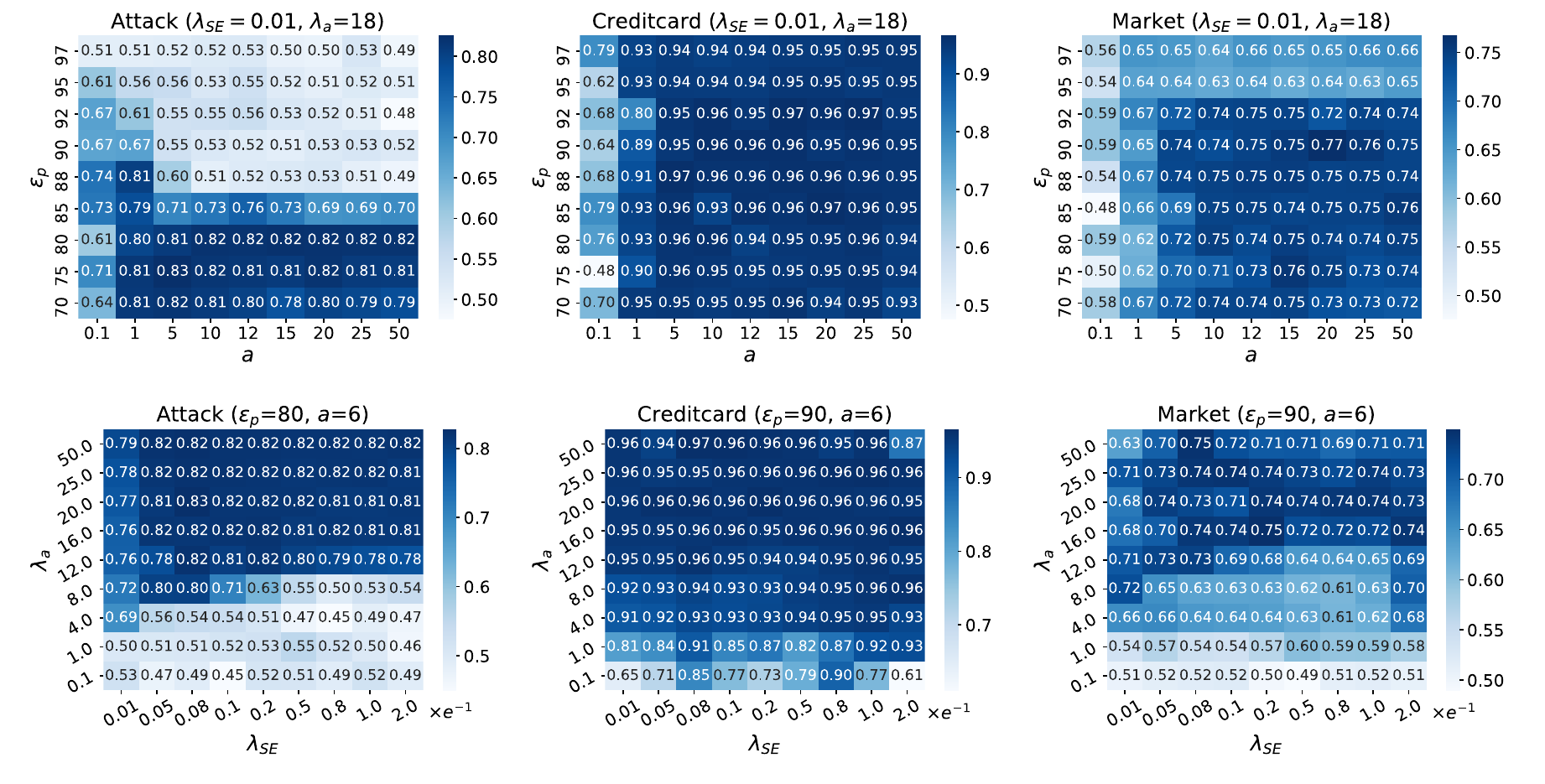}
        \end{center}
        \caption{The AUC-ROC values for SG-AE with various sets of parameters on three datasets, \textit{attack}, \textit{credictcard}, and \textit{martket}. We first fix $\lambda_{SE}=0.01$ and $\lambda_a=18$ for each dataset and train the SG-AE with different values of $\epsilon_p$ and $a$. Then we fix $\epsilon_p$ and $a$, and train the SG-AE with different values of $\lambda_{SE}$ and $\lambda_a$.}
        \label{fig:heatmap}
    \end{figure*}
    
    Figure~\ref{fig:heatmap} reports the AUC-ROC values of the two-step testing on three datasets, \textit{attack}, \textit{creditcard}, and \textit{market}. The darker colors suggest better performance. We note that state-of-the-art results can be achieved without complicated and time-consuming hyperparameter searching. In addition, we find that 
    (i) $\epsilon_p$ and $\lambda_{a}$ have relatively greater impact on model performances than $a$ and $\lambda_{SE}$.
    (ii) Although dataset \textit{attack} (25.24\%), \textit{credictcard} (0.17\%), and \textit{martket} (11.70\%) have different anomaly rates, SG-AE can achieve optimal performances with threshold $\epsilon_p$ smaller than 80\%, suggesting that our score-guidance strategy does not require the exact anomaly rate and can be effective in unsupervised scenarios. Moreover, we can utilize the variations of the score distribution under different $\epsilon_p$ to approximately observe the true anomaly rate. 
    (iii) The hyperparameter $a$ determines the target for the guided scores of abnormal samples and the performances of SG-AE change lightly when $a>5$. The results reflect that a too-large $a$ only extends the distribution of scores without further performance improvements. In addition, the performances also remain stable with variations of $\lambda_{SE}$, indicating that we do not need to pay efforts to tune these two parameters.
    (iv) $\lambda_a$ adjusts the loss effect of the suspected abnormal part and a larger $\lambda_a$ is preferred, revealing the difficulty of distinguishing those non-obvious normal samples in the transition fields. 
    In summary, although the scoring network introduces four parameters, the score-guided model is not much sensitive to these parameters. When the scoring network transfers to a new model or an SG-Model applies to a new dataset, $\epsilon_p$ and $\lambda_{a}$ are the core parameters to tune and the score distribution will help the tuning process.

\subsubsection{Robustness to Different Anomaly Rates}

    As we intend to learn the disparity between the normal and abnormal data using a scoring network, we can expect that the performance of SG-AE would be robust to the ratio of the abnormal data. To validate this expectation, we reorganize the two datasets with high anomaly rates, \textit{attack} and \textit{intrusion}, and compare SG-AE with iForest, AE, RDP, and ECOD on these datasets with different anomaly rates. Specifically, we randomly drop anomaly samples in the training set, adjusting the anomaly rate to $[5\%, 10\%, 15\%, 20\%, 25\%]$ on \textit{attack} and $[5\%, 10\%, 15\%, 20\%, 25\%, 30\%, 35\%]$ on \textit{intrusion}. The validation and testing sets keep the original anomaly rate.
    
\begin{figure}[htb!]
    \begin{center}
    \includegraphics[width=0.48\textwidth]{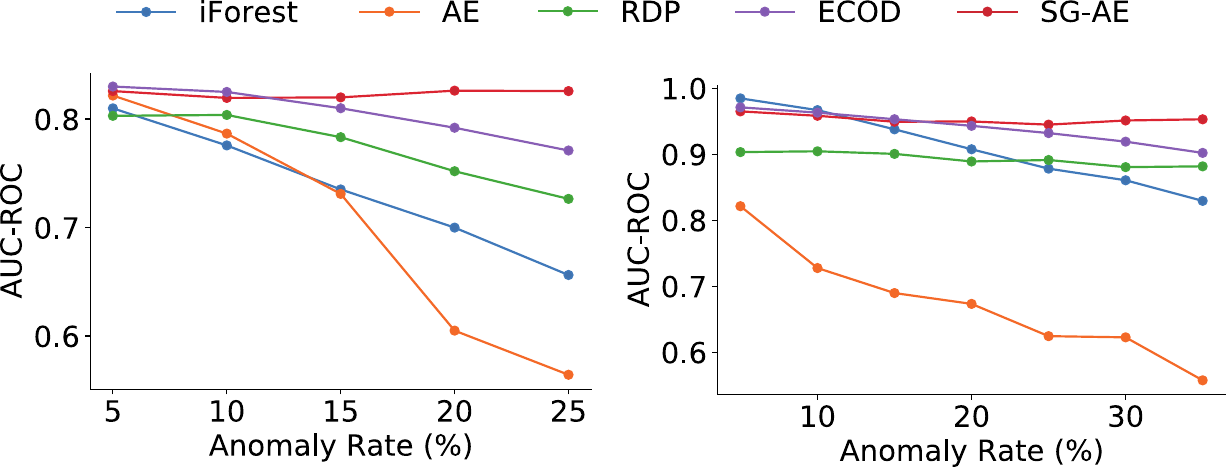}
    \end{center}
    \caption{The AUC-ROC values of selected models in different anomaly rates on two datasets \textit{attack} and \textit{intrusion}. SG-AE achieves stable AUC-ROC with various anomaly rates.}
    \label{fig:rate}
    \end{figure}

    Figure~\ref{fig:rate} presents the comparison results in different anomaly rates. Obviously, SG-AE achieves a more stable performance than other competing models, while AE has the fastest AUC-ROC performance degradation on both datasets. The strong sensitivity of AE to the anomaly rate is mainly due to its intention to reconstruct all input data, while a large fraction of abnormal data hinders the training of the reconstruction model. We also observe that the AUC-ROC values of iForest and RDP drop faster in \textit{attack} than in \textit{intrusion}, suggesting that the robustness to anomaly rates also depends on the characteristic of data.
    
\subsection{Extension to SOTA models}
    
    To examine the transferability of the scoring network to other models, we evaluate the performances of four extended models in the seven tabular datasets.   
    
    \textbf{Settings}. We extend the scoring network to four state-of-the-art models: DAGMM, RDA, RSRAE, and GOAD. The extended models are named score-guided DAGMM (SG-DAGMM), score-guided RDA (SG-RDA), score-guided RSRAE (SG-RSRAE) and score-guided GOAD (SG-GOAD). 
    For the input of the scoring network, SG-DAGMM takes the same input as the compression network in DAGMM, SG-RDA takes the output of the encoder, SG-RSRAE takes the output of the RSR layer, and SG-GOAD takes the output of the representation learner. Here the datasets are configured the same as the experiments of SG-AE. We also search parameters to adapt the SG-Models to these datasets. Specifically, SG-RDA is tuned with its original $\lambda$ and $\lambda_{\rm{SE}}$, SG-DAGMM and SG-RSRAE are tuned with their two original $\lambda$ and $\lambda_{\rm{SE}}$, and SG-GOAD is tuned with $\lambda_{\rm{SE}}$ and $\lambda_{\rm{a}}$.
    
    \textbf{Performance evaluation}. The results of AUC-ROC and AUC-PR are shown in Table~\ref{tab:sg-sota}. Compared with the performances of the original models as shown in Table~\ref{tab:performance}, these SG-Models achieve varying degrees of improvement in most datasets. 
    Specifically, the averaged improvements of SG-AE, SG-RDA, SG-RSRAE, SG-DAGMM, and SG-GOAD on AUC-ROC than the original models on the seven datasets are 20.5\%, 15.1\%, 14.1\%, 10.6\%, and 2.3\%, while the averaged improvements on AUC-PR are 45.2\%, 27.7\%, 63.6\%, 170.2\%, and 6.9\%.
    These results suggest that the score-guidance strategy can better handle the contaminated data. In addition, SG-AE and SG-DAGMM obtain more obvious performance improvements and SG-DAGMM even surpasses SG-AE on some datasets. This is mainly because the scoring network enables them to deal with anomaly contamination and generates better data representation. 
    Although GOAD has demonstrated some improvements in performance, these improvements are relatively modest compared to other baseline methods. This can be attributed to two factors. Firstly, GOAD's initial performance is already strong, thereby limiting the potential for further improvement. Secondly, the structure of GOAD's representation learner is relatively simple, which also constrains the potential for optimization.
    In summary, these experimental results not only demonstrate the effectiveness of the scoring network design but also reflect that the scoring network can be incorporated into different unsupervised anomaly detection models.

    \begin{table*}[hbt]
    \caption{AUC-ROC and AUC-PR (mean$\pm$std) of SG-SOTA models. (*) indicates the performance improvement of SG-SOTA models over the original models.}
    \label{tab:sg-sota}
    \centering
    \scalebox{0.86}{
        \setlength\tabcolsep{5pt}
        \begin{tabular}{c|l|lllll}
        \toprule
            & \textbf{Datasets} & \textbf{SG-AE} & \textbf{SG-RDA} &   \textbf{SG-RSRAE} & \textbf{SG-DAGMM} & \textbf{SG-GOAD} \\
        \midrule 
            \multirow{7}{*}{\textbf{AUC-ROC}} & 
            Attack      & \textbf{0.833$\pm$0.009} (48.5\%) & 0.816$\pm$0.016 (32.7\%)    & 0.771$\pm$0.064 (13.2\%)  & 0.783$\pm$0.044 (2.8\%)  & 0.785$\pm$0.010 (7.5\%) \\   
            & Bcsc        & \textbf{0.912$\pm$0.008} (3.9\%)  & 0.905$\pm$0.004 (3.8\%)   & 0.868$\pm$0.043 (3.3\%)  & 0.785$\pm$0.117 (16.8\%) & 0.898$\pm$0.012 (1.8\%)  \\                
            & Creditcard  & \textbf{0.964$\pm$0.004} (6.8\%) & 0.940$\pm$0.010 (2.2\%)   & 0.930$\pm$0.019 (-1.3\%)  & 0.915$\pm$0.021 (9.8\%)  & 0.941$\pm$0.015 (0.2\%)  \\        
            & Diabetic    & 0.548$\pm$0.007 (2.4\%) & 0.561$\pm$0.027 (4.3\%)  & 0.545$\pm$0.012 (14.3\%)     & \textbf{0.581$\pm$0.066} (13.0\%)  & 0.514$\pm$0.016 (-1.9\%) \\
            & Donor       & 0.541$\pm$0.004 (11.3\%) & 0.540$\pm$0.007 (5.3\%)   & \textbf{0.597$\pm$0.059} (23.3\%) & 0.568$\pm$0.049 (9.4\%)    & 0.508$\pm$0.013 (2.6\%) \\                   
            & Intrusion   & 0.906$\pm$0.016 (59.5\%) & 0.899$\pm$0.019 (46.4\%)   & 0.808$\pm$0.144 (9.0\%) & \textbf{0.976$\pm$0.003} (13.5\%)  & 0.916$\pm$0.017 (2.0\%) \\            
            & Market      & \textbf{0.750$\pm$0.025} (10.8\%) & 0.747$\pm$0.033 (10.8\%)   & 0.733$\pm$0.030 (36.8\%)  & 0.698$\pm$0.003 (9.1\%)     & 0.737$\pm$0.018 (3.9\%) \\       
        \midrule
            \multirow{7}{*}{\textbf{AUC-PR}} &
            Attack & 0.596$\pm$0.024 (96.7\%) & 0.534$\pm$0.027 (66.4\%)  & 0.529$\pm$0.096 (-0.8\%)   & \textbf{0.660$\pm$0.035} (44.4\%) & 0.454$\pm$0.013 (12.4\%) \\
            & Bcsc & \textbf{0.816$\pm$0.004} (32.3\%) & 0.805$\pm$0.001 (31.3\%)  & 0.510$\pm$0.219 (272.3\%)   & 0.421$\pm$0.380 (694.3\%) & 0.640$\pm$0.017 (6.1\%) \\
            & Creditcard  & 0.291$\pm$0.072 (38.6\%) & 0.202$\pm$0.083 (-7.3\%)     & 0.222$\pm$0.100 (40.5\%)   & \textbf{0.494$\pm$0.169} (394.0\%)  & 0.144$\pm$0.015 (25.2\%) \\
            & Diabetic   & 0.142$\pm$0.004 (8.4\%) & 0.143$\pm$0.013 (8.3\%)   & 0.135$\pm$0.005 (15.4\%)     & \textbf{0.168$\pm$0.049} (31.3\%) & 0.118$\pm$0.006 (-6.3\%) \\
            & Donor   & 0.075$\pm$0.002 (19.0\%) & 0.078$\pm$0.003 (18.2\%)   & \textbf{0.081$\pm$0.011} (35.0\%) & 0.076$\pm$0.005 (10.1\%) & 0.063$\pm$0.003 (0.0\%) \\
            & Intrusion  & 0.904$\pm$0.035 (96.1\%) & 0.758$\pm$0.030 (58.6\%)   & 0.736$\pm$0.188 (-1.5\%) & \textbf{0.971$\pm$0.003} (21.2\%)  & 0.766$\pm$0.012 (7.4\%) \\
            & Market   & \textbf{0.270$\pm$0.030} (25.6\%) & 0.251$\pm$0.031 (19.5\%)   & 0.238$\pm$0.033 (84.5\%)  & 0.203$\pm$0.011 (-3.8\%)  & 0.232$\pm$0.018 (3.6\%) \\  
        \bottomrule
        \end{tabular}
    }
    \end{table*}

\subsection{Extension to document and image tasks}
    
    To examine the transferability to different tasks, we compare SG-AE with iForest, DAGMM, RDA, RSRAE, and RDP on document and image datasets. 
    
    \textbf{Settings}. For the document task, we utilize datasets, \textit{news} and \textit{reuters}~\cite{rsrae}. \textit{News} involves newsgroup documents with 20 different labels and \textit{reuters} contains 5 text categories. Following the preprocessing steps in~\cite{rsrae}, \textit{news} and \textit{reuters} are randomly split into 360 documents per class and the documents are embedded into vectors of size $10,000$ and $26,147$, respectively. In each round of testing, we choose one class as normal data in turns and randomly select abnormal samples from other classes. By collecting different numbers of abnormal data, we compare the SG-AE with baselines in the document datasets with varying anomaly rates, $[5\%, 10\%, 15\%, 20\%, 25\%]$. 
    For image task, we take \textit{mnist} dataset as an example~\cite{rda}. \textit{mnist} consists of 5124 instances, with the anomaly rate of $5.2\%$. The normal data are the image of digit "4" and the abnormal data are other digits. For each task, we average the AUC-ROC and AUC-PR values over 10 independent runs. The training and testing sets are 8/2. The layer sizes of the autoencoder framework are [1024, 256, 64, 20] in document tasks and [128, 64, 32] in the image task. The batch size is set as 32 and other settings are consistent with the previous experiments. We also conduct hyperparameter searching to achieve nearly optimal results. 
    
    \begin{figure}[htb!]
        \begin{center}
        \includegraphics[width=0.46\textwidth]{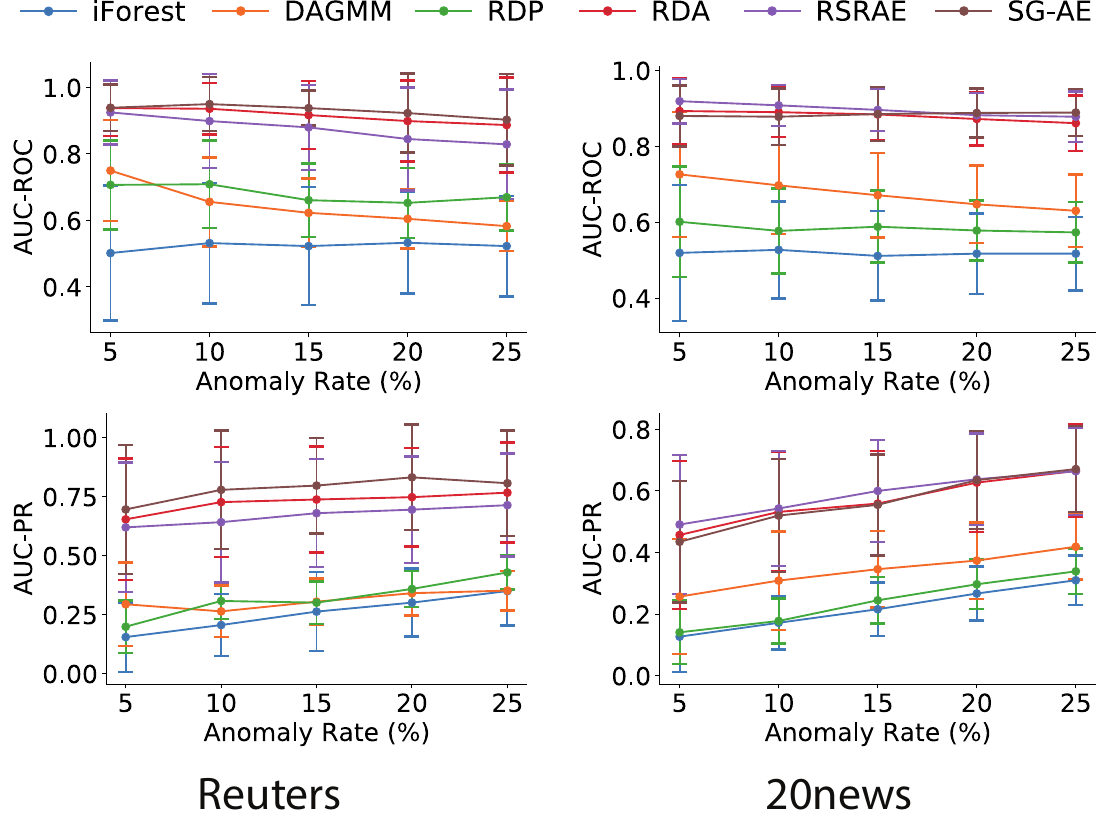}
        \end{center}
        \caption{AUC-ROC and AUC-PR values on document data.}
        \label{fig:nlp}
    \end{figure}

    \textbf{Performance evaluation}. 
    Figure~\ref{fig:nlp} illustrates the results for document datasets. SG-AE achieves the best performance in \textit{reuters}. In \textit{20news}, SG-AE performs slightly weaker than RSRAE and RDA at a smaller anomaly rate but performs better at a larger anomaly rate. 
    This might be due to the different granularity when handling anomalies. Specifically, 
    RSRAE and RDA filter abnormal parts for each sample at the feature level, which is effective for datasets with large feature correlations (such as document or image datasets).
    SG-AE simply guides the sample distribution at the data level. In fact, the feature-level and the data-level strategies are not in conflict. The combination of these two strategies can be further studied in the future.
    The results for the image dataset are shown in Table~\ref{tab:mnist}. Similar to the document datasets, SG-AE has competitive results on AUC-ROC but performs weaker than RSRAE in terms of AUC-PR. 
    Therefore, we further evaluate SG-RSRAE, achieving performance improvements of $1.3\%$ on AUC-ROC and $19.7\%$ on AUC-PR over RSRAE. 
    These results confirm the effectiveness of the score-guidance strategy in different tasks.

    \begin{table}[htb!]
    \caption{AUC-ROC and AUC-PR values on \textit{mnist} dataset.}
        \label{tab:mnist}
        \centering
        \begin{tabular}{l|cc}
        \toprule
             & \textbf{AUC-ROC} & \textbf{AUC-PR}   \\
        \midrule
            \textbf{iForest} & 0.873$\pm$0.014 & 0.372$\pm$0.036 \\
            \textbf{DAGMM}   & 0.758$\pm$0.040 & 0.235$\pm$0.036	\\
            \textbf{RDP}     & 0.888$\pm$0.024 & 0.475$\pm$0.065	\\
            \textbf{RDA}	    & 0.912$\pm$0.010 & 0.556$\pm$0.038	\\
            \textbf{RSRAE}	& 0.939$\pm$0.007 & 0.615$\pm$0.026 \\	
            \textbf{SG-AE}	& 0.939$\pm$0.005 & 0.563$\pm$0.015	\\
            \textbf{SG-RSRAE}	& \textbf{0.951$\pm$0.010} & \textbf{0.736$\pm$0.062}	\\
        \bottomrule
        \end{tabular}
    \end{table}

\section{Discussion on Hyperparameter Selection}
\label{sec:param}

    Hyperparameter selection is a common challenge for unsupervised deep learning methods, especially when we don't have a validation dataset. 
    Our proposed method contains four hyperparameters. They are, obvious-normal percentile $\epsilon_p$, target anomaly score $a$, and two loss weights $\lambda_{\rm{SE}}$ and $\lambda_{\rm{a}}$, which commonly appear in multi-task learning. 
    The hyperparameter $\epsilon_p$ is highly related to the anomaly ratio of a specific dataset, while the rest can be set to fixed values because the method is not so sensitive to them, as discussed in the experiment. 
    In the real-world scenario, experts can easily set $\epsilon_p$ according to their experience in the task. If we turn to a new dataset without knowing the concept of abnormal samples, we suggest to use the standardized reference scores $\hat{f}$ to determine $\epsilon_p$, which is formulated as follows: 
    \begin{equation}\label{eq:ep}
        \epsilon_{p} = \frac{N(\hat{f} < std(\hat{f}))}{N(\hat{f})}, \hat{f} = \frac{f-mean(f)}{std(f)}, 
    \end{equation}
    where $f$ is the reference score introduced in section~\ref{sec:SGR}, $mean(*)$ and $std(*)$ calculate the mean and standard deviation of all $*$, and $N(*)$ counts the number of $*$. The hyperparameter $\epsilon_p$ is updated with Eq~(\ref{eq:ep}) every training epoch.

    We conduct experiments to evaluate the effectiveness of Eq~(\ref{eq:ep}) on five tabular datasets. The hyperparameters $a$, $\lambda_{\rm{SE}}$, and $\lambda_{\rm{a}}$ are set to fixed values of 10, 0.01, and 20. The SG-AEs with fixed $\epsilon_p=90$ and $\epsilon_p=80$ are denoted as SG-AE$_{90}$ and SG-AE$_{80}$. The SG-AE with Eq~(\ref{eq:ep}) to determine $\epsilon_p$ is denoted as SG-AE$\rm{_{ep}}$. Besides, we propose a variant SG-AE$\rm{_{epa}}$, which is based on SG-AE$\rm{_{ep}}$ but uses $tanh()$ to constrain the anomaly scores to $(-1,1)$. This variant discards the hyperparameter $a$ and the target anomaly score is $1$. We also include the tuned SG-AE, denoted as SG-AE$\rm{_{tuned}}$, and ECOD~\cite{li2022ecod} in comparison.  

    \begin{table}[htb!]
    \caption{Performance Comparisons in AUC-ROC and AUC-PR.}
        \label{tab:hyper}
        \centering
        \scalebox{0.84}{
            \setlength\tabcolsep{5pt}
            \begin{tabular}{cl|ccccc}
            \toprule
                & & \textbf{Attack} & \textbf{Bcsc} & \textbf{Creditcard}
                & \textbf{Intrusion} & \textbf{Market} \\
            \midrule
                \multirow{5}{*}{\textbf{AUC-ROC}} 
                & \textbf{ECOD}  & 0.770  & 0.749  & 0.934  & \underline{0.903}  & 0.676  \\
                & \textbf{SG-AE}\boldmath$\rm{_{90}}$  & 0.547  & 0.855  & 0.926  & 0.561  & \underline{0.741}  \\
                & \textbf{SG-AE}\boldmath$\rm{_{80}}$  & \underline{0.831}  & 0.832  & 0.916  & 0.845  & 0.729  \\
                & \textbf{SG-AE}\boldmath$\rm{_{ep}}$  & 0.819  & \underline{0.877}  & \underline{0.944}  & 0.866  & \underline{0.741}  \\
                & \textbf{SG-AE}\boldmath$\rm{_{epa}}$  & 0.827  & 0.853  & 0.939  & 0.841  & 0.666  \\
                & \textbf{SG-AE}\boldmath$\rm{_{tuned}}$  & \textbf{0.833}  & \textbf{0.912}  & \textbf{0.964}  & \textbf{0.906}  & \textbf{0.750}  \\
            \midrule
                \multirow{5}{*}{\textbf{AUC-PR}} 
                & \textbf{ECOD}  & 0.457  & 0.075  & \underline{0.161}  & \underline{0.762}  & 0.215  \\
                & \textbf{SG-AE}\boldmath$\rm{_{90}}$  & 0.283  & 0.294  & 0.054  & 0.429  & \underline{0.244}  \\
                & \textbf{SG-AE}\boldmath$\rm{_{80}}$  & \underline{0.569}  & 0.321  & 0.071  & 0.729  & 0.229  \\
                & \textbf{SG-AE}\boldmath$\rm{_{ep}}$  & 0.538  & \underline{0.475}  & 0.085  & 0.745  & 0.240  \\
                & \textbf{SG-AE}\boldmath$\rm{_{epa}}$  & 0.564  & 0.369  & 0.061  & 0.702  & 0.200  \\
                & \textbf{SG-AE}\boldmath$\rm{_{tuned}}$  & \textbf{0.596}  & \textbf{0.816}  & \textbf{0.291}  & \textbf{0.904}  & \textbf{0.270}  \\
                
            \bottomrule
            \end{tabular}
        }
    \end{table}

    The results are the average of ten independent runs and are shown in Table~\ref{tab:hyper}, where the best and the second-best results are highlighted in bold and underline respectively. 
    (1) Compared with ECOD, SG-AE$_{90}$ performs better on \textit{bcsc} and \textit{market} but worse on the other three datasets. When $\epsilon_p$ turns to 80, SG-AE$_{80}$ achieves significant performance improvements on \textit{attack} and \textit{intrusion}, because $\epsilon_p=80$ is closer to the true normal ratios of these two datasets. This indicates that $\epsilon_p$ has a huge impact on performance, especially when it is much smaller than the true normal ratio. 
    (2) SG-AE$_{ep}$ obtains the second-best results on three datasets and outperforms ECOD on four datasets in AUC-ROC, which demonstrates the effectiveness of the proposed Eq~(\ref{eq:ep}). It is because the automatically set $\epsilon_p$ utilizes the distribution of the reference scores on the dataset. This strategy is adaptable to different datasets and has fewer wrong guidances than manually setting an unmatched $\epsilon_p$.
    (3) SG-AE$_{epa}$ performs worse than SG-AE$_{ep}$ overall because constraining the score to (-1,1) requires the model to be more sensitive to numerical differences, making it more difficult to distinguish anomalies.
    (4) SG-AE$_{tuned}$ performs best on the five datasets, indicating that flexible hyperparameter settings contribute to a higher level of performance.  
    In summary, we can use Eq~(\ref{eq:ep}) to automatically set $\epsilon_p$ with fixed three other hyperparameters, when we apply this method to a brand new dataset. If we have a validation dataset, we can tune the model to achieve better performances.

\section{Conclusion and future work} 
    Targeting the unsupervised anomaly detection task, this work introduced an effective scoring network with score-guided regularization.
    The proposed scoring network can be integrated into existing deep unsupervised representation learning-based anomaly detection methods and enhances their anomaly detection capabilities. Specifically, the scoring network utilizes the original discrimination metric as a self-supervised signal to pre-judge the obvious-normal samples and leverage these samples to strengthen the representation learner and enlarge the score disparity between obvious-normal and suspected abnormal samples. Moreover, the scoring network learns the discrimination metric according to the dataset and directly outputs the anomaly scores, enhancing the transferability of the original method. 
    We next proposed a representative instantiation of incorporating the scoring network into an autoencoder framework, namely score-guided autoencoder (SG-AE). We first conducted experiments on synthetic datasets to examine the effectiveness of the score-guidance strategy. The results show that the anomaly score disparity between normal and abnormal data continues to expand during training, especially in the transition field.  
    Then comprehensive experiments are conducted on seven tabular datasets, suggesting that the proposed SG-AE is competitive with the state-of-the-art methods in terms of both AUC values and the Kolmogorov–Smirnov statistical index. 
    We also analyzed the sensitivity of the introduced parameters in our scoring network and found that different datasets can share similar configurations.
    In addition, as the scoring network can learn the disparity between normal and abnormal data, we expect that SG-AE can work well even with a large fraction of anomaly data. To validate this expectation, we then tested the SG-AE on two datasets with varying anomaly rates, indicating that SG-AE is more robust to the anomaly rate than the other three baselines.
    In order to present the transferability of the scoring network on different unsupervised representation learning-based methods, we then applied the scoring network to four state-of-the-art methods and the experimental results approve the performance improvement.
    Moreover, we applied the scoring network to two document datasets and one image dataset to present the transferability to different anomaly detection tasks. Experimental results show that SG-Models are comparable to the state-of-the-art methods, especially for datasets with large anomaly rates.
    Lastly, we further discussed the hyperparameter selection and proposed a method to automatically set $\epsilon_p$, which is highly related to datasets. We also conducted experiments to demonstrate its effectiveness.

    There are several improvements and potential extensions that merit further study: 
    (1) Adjusting the $\lambda$ parameters to balance the effects of several loss function terms is challenging when applying the scoring network to a method with multiple loss functions. Potential solutions could be utilizing the multi-objective optimization techniques~\cite{konakovic2020diversity,liu2021towards}, which attempt to balance the trade-offs between several objectives. Hyperparameter tuning techniques~\cite{rln,mackay2019self}, which map hyperparameters into loss functions and optimize them during training, are also worthy of consideration to deal with the problem of too many hyperparameters.  
    (2) The score-guided strategy deserves to be further applied to convolutional frameworks or sequential frameworks to evaluate the performance changes in various anomaly detection tasks with image, time-series, or graph datasets.

\ifCLASSOPTIONcompsoc
  \section*{Acknowledgments}
\else
  \section*{Acknowledgment}
\fi

This work was jointly supported by the National Natural Science Foundation of China (62102258), Shanghai Pujiang Program (21PJ1407300), Shanghai Municipal Science and Technology Major Project (2021SHZDZX0102), Shanghai Key Laboratory of Multidimensional Information Processing, East China Normal University (MIP20225) and the Fundamental Research Funds for the Central Universities.

\ifCLASSOPTIONcaptionsoff
  \newpage
\fi

\bibliographystyle{IEEEtran}
\bibliography{citationlist}

\begin{IEEEbiography}[{\includegraphics[width=1in,height=1.25in,clip,keepaspectratio]{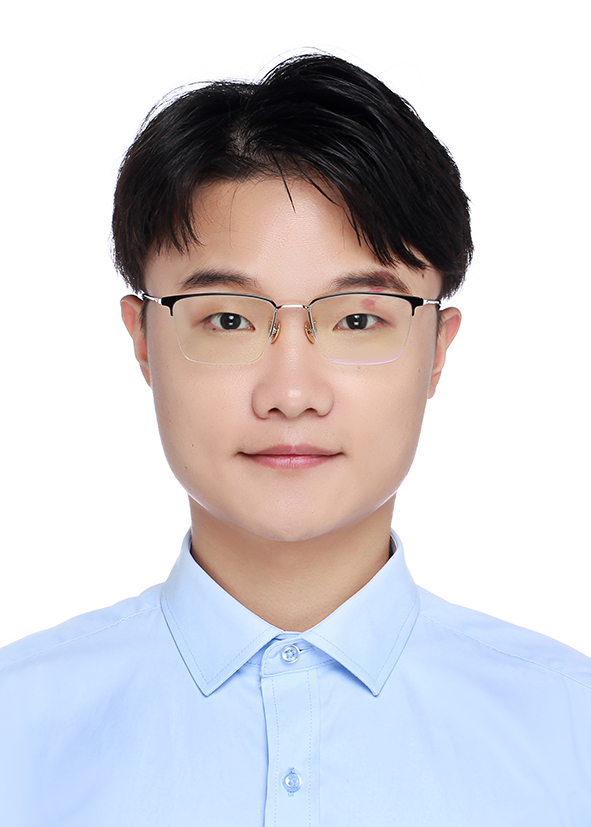}}]{Zongyuan Huang,} received the B.E. degree from Huazhong University of Science and Technology in 2019, is currently working toward the Ph.D. degree in the Department of Electronic Engineering, Shanghai Jiao Tong University, China. His research interests lie in machine learning and data mining with particular focuses on anomaly detection, urban computing, and mobility data mining. 
\end{IEEEbiography}

\vspace{-10 mm} 
\begin{IEEEbiography}[{\includegraphics[width=1in,height=1.25in,clip,keepaspectratio]{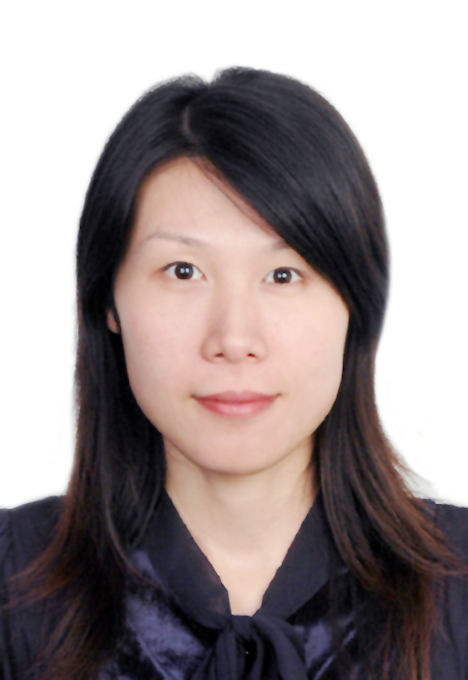}}]{Baohua Zhang} received her Doctor's Degree in Computer Software and Theory from Fudan University in 2009. She joined the Software Development Center of ICBC in 2009 and is currently working in the Big Data and AI Lab of ICBC as a technical manager. She is now the team leader of big data and AI technology scouting in the financial field.
\end{IEEEbiography}

\vspace{-10 mm} 
\begin{IEEEbiography}[{\includegraphics[width=1in,height=1.25in,clip,keepaspectratio]{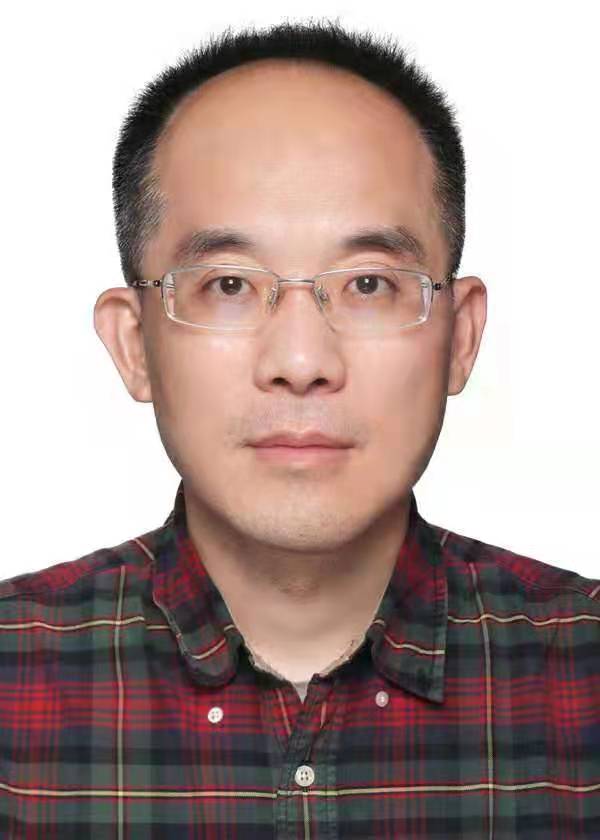}}]{Guoqiang Hu} received his M.Sc. and Dr.-Ing. Degrees in Electrical Engineering from the University of Stuttgart, Germany in 2002 and 2007, respectively. From 2008 to 2010, he was a research fellow in the Norwegian University of Science and Technology (NTNU), focusing on the domain of network architecture design and quality-of-service provisioning. From 2010 to 2021, he was a research staff member in IBM China Research Lab and led innovation projects in the field of Internet-of-Things and computer vision. He joined the Big Data and AI Lab of ICBC in 2021 and is the technical director of technology scouting and computer vision.
\end{IEEEbiography}

\vspace{-10 mm} 
\begin{IEEEbiography}[{\includegraphics[width=1in,height=1.25in,clip,keepaspectratio]{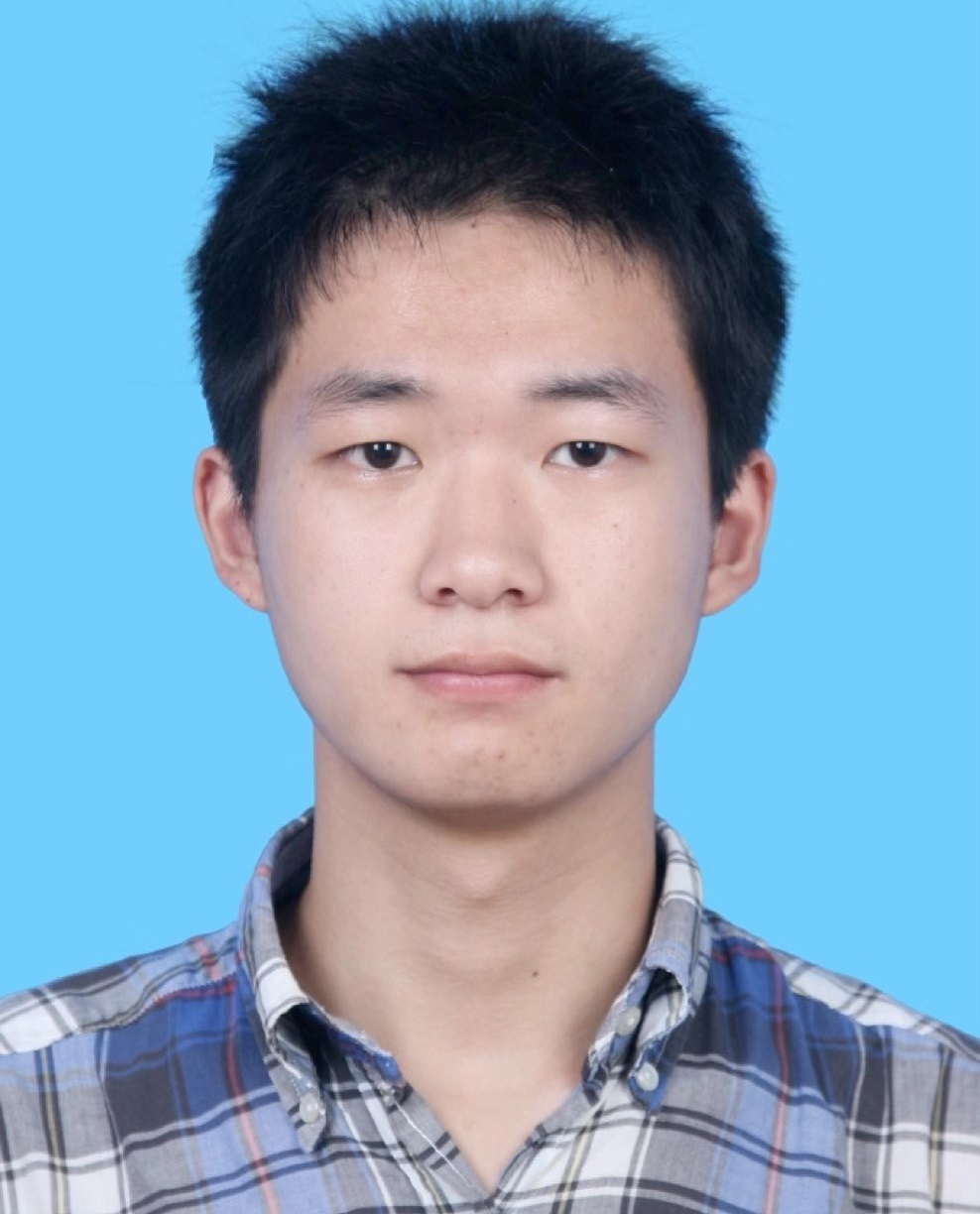}}]{Longyuan Li} received the B.E. degree in Electronic Engineering from Huazhong University of Science and Technology in 2015. He is currently working toward the Ph.D. degree in the Department of Electronic Engineering, Shanghai Jiao Tong University, China. His research interests include machine learning and data mining. He focuses on probabilistic models and Bayesian non-parametric models, for sequential data such as multi-dimensional time-series. He is also interested in deep probabilistic graphical models and uncertainty estimation, with first-authored publications in IJCAI'19, AAAI'21, IEEE TNNLS.
\end{IEEEbiography}

\vspace{-10 mm} 
\begin{IEEEbiography}[{\includegraphics[width=1in,height=1.25in,clip,keepaspectratio]{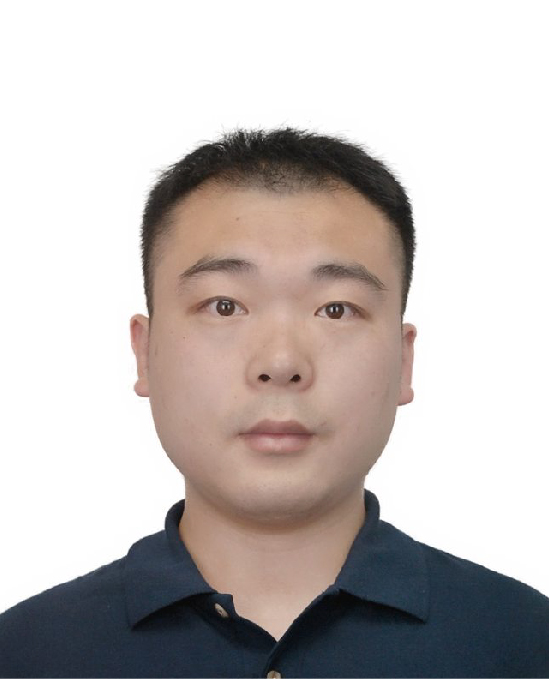}}]{Yanyan Xu,} received the Ph.D. degree from the Department of Automation, Shanghai Jiao Tong University in 2015. He is now an associate professor in the AI Institute, Shanghai Jiao Tong University. Prior to joining SJTU, he was a postdoctoral associate in the Human Mobility and Networks (HuMNet) Lab in the Department of Urban and Regional Planning at UC Berkeley and the Department of Civil and Environmental Engineering at MIT from 2015 to 2020. He was also a guest postdoctoral fellow in the Energy Analysis and Environmental Impacts Division, Lawrence Berkeley National Laboratory from 2017 to 2018. His research lies in AI for complex systems, human mobility and urban computing, with particular emphasis placed on the use of massive trajectory data in Urban Science, Transportation, Energy, and Environment, from interdisciplinary perspectives. His work has been published in Nature Energy, Nature Computational Science, Science Advances, J. Roy. Soc. Interface, IEEE Trans. ITS, TRC, and IJCAI, among others.
\end{IEEEbiography}

\vspace{-10 mm} 
\begin{IEEEbiography}[{\includegraphics[width=1in,height=1.25in,clip,keepaspectratio]{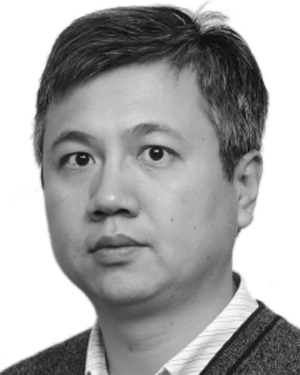}}]{Yaohui Jin} was once a Technical Staff Member with Bell Labs Research China. After that he joined Shanghai Jiao Tong University in 2002, where he is a Professor with the State Key Laboratory of Advanced Optical Communication Systems and Networks and the Deputy Director of Network and Information Center. His research interests include civic engagement and open innovation, cloud computing network architecture, and streaming data analysis. He is the Founder of OMNILab, which is an open innovation lab focusing on data analysis. He has published over 100 technical papers in leading conferences and journals and is the owner of over 10 patents. In 2014, OMNILab won the champion of CCF national big data challenge among nearly 1000 teams, and was the champion of the Shanghai open data innovation and creation competition. He has served over 10 technical committees. He enthuses public service and science popularization, actively promotes crowd engaged innovation, and interdisciplinary collaboration.
\end{IEEEbiography}

\end{document}